\documentclass[11pt]{article}

\usepackage[a4paper,margin=1in]{geometry}
\usepackage[T1]{fontenc}
\usepackage{lmodern}
\usepackage{microtype}
\usepackage{url}
\usepackage{hyperref}

\usepackage{algorithm}
\usepackage{algorithmic}
\usepackage{booktabs}
\usepackage{makecell}
\usepackage{listings}
\usepackage{tikz}
\usetikzlibrary{arrows.meta,calc,positioning}
\providecommand{\Description}[1]{}

\usepackage[numbers,sort&compress]{natbib}

\definecolor{heathi}{RGB}{8,48,107}

\title{Universal Multi-Modal Traceformer: Integrating Heterogeneous Context for Process Event Prediction}

\author{
  Fabian Spaeh\\
  Celonis\\
  \texttt{f.spaeh@celonis.com}
  \and
  Jingxing Fang\\
  Celonis\\
  \texttt{j.fang@celonis.com}
  \and
  Shandian Zhe\\
  Celonis\\
  \texttt{s.zhe@celonis.com}
  \and
  Bin Shen\\
  Celonis\\
  \texttt{b.shen@celonis.com}
}

\usepackage{amsmath,amsfonts}

\begin{document}

% Math commands by Thomas Minka
\newcommand{\var}{{\rm var}}
\newcommand{\vtrans}[2]{{#1}^{(#2)}}
\newcommand{\kron}{\otimes}
\newcommand{\schur}[2]{({#1} | {#2})}
\newcommand{\schurdet}[2]{\left| ({#1} | {#2}) \right|}
\newcommand{\had}{\circ}
\newcommand{\diag}{{\rm diag}}
\newcommand{\invdiag}{\diag^{-1}}
\newcommand{\rank}{{\rm rank}}
 \newcommand{\expt}[1]{\langle #1 \rangle}
% careful: ``null'' is already a latex command
\newcommand{\nullsp}{{\rm null}}
\newcommand{\tr}{{\rm tr}}
\renewcommand{\vec}{{\rm vec}}
\newcommand{\vech}{{\rm vech}}
\renewcommand{\det}[1]{\left| #1 \right|}
\newcommand{\pdet}[1]{\left| #1 \right|_{+}}
\newcommand{\pinv}[1]{#1^{+}}
\newcommand{\erf}{{\rm erf}}
\newcommand{\hypergeom}[2]{{}_{#1}F_{#2}}
\newcommand{\mcal}[1]{\mathcal{#1}}
\newcommand{\bepsilon}{\boldsymbol{\epsilon}}
\newcommand{\brho}{\boldsymbol{\rho}}
% boldface characters
\renewcommand{\a}{{\bf a}}
\renewcommand{\b}{{\bf b}}
\renewcommand{\c}{{\bf c}}
\renewcommand{\d}{{\rm d}}  % for derivatives
\newcommand{\e}{{\bf e}}
\newcommand{\f}{{\bf f}}
\newcommand{\g}{{\bf g}}
\newcommand{\h}{{\bf h}}
\newcommand{\bi}{{\bf i}}
\newcommand{\bj}{{\bf j}} 

%\newcommand{\k}{{\bf k}}
% in Latex2e this must be renewcommand
\renewcommand{\k}{{\bf k}}
\newcommand{\m}{{\bf m}}
\newcommand{\mhat}{{\overline{m}}}
\newcommand{\tm}{{\tilde{m}}}
\newcommand{\n}{{\bf n}}
\renewcommand{\o}{{\bf o}}
\newcommand{\p}{{\bf p}}
\newcommand{\q}{{\bf q}}
\newcommand{\wy}{{\widehat{\y}}}
\newcommand{\wlam}{{\widehat{\lambda}}}
\renewcommand{\r}{{\bf r}}
\newcommand{\s}{{\bf s}}
\renewcommand{\t}{{\bf t}}
\renewcommand{\u}{{\bf u}}
\renewcommand{\v}{{\bf v}}
\newcommand{\w}{{\bf w}}
\newcommand{\x}{{\bf x}}
\newcommand{\y}{{\bf y}}
\newcommand{\z}{{\bf z}}
\newcommand{\A}{{\bf A}}
\newcommand{\B}{{\bf B}}
\newcommand{\C}{{\bf C}}
\newcommand{\D}{{\bf D}}
\newcommand{\F}{{\bf F}}
\newcommand{\G}{{\bf G}}
\newcommand{\Gcal}{{\mathcal{G}}}
\newcommand{\Dcal}{\mathcal{D}}
\newcommand{\Qcal}{{\mathcal{Q}}}
\newcommand{\Pcal}{{\mathcal{P}}}
\newcommand{\Hcal}{{\mathcal{H}}}
\renewcommand{\H}{{\bf H}}
\newcommand{\I}{{\bf I}}
\newcommand{\J}{{\bf J}}
\newcommand{\K}{{\bf K}}
\renewcommand{\L}{{\bf L}}
\newcommand{\Lcal}{{\mathcal{L}}}
\newcommand{\M}{{\bf M}}
\newcommand{\Mcal}{{\mathcal{M}}}
\newcommand{\Ocal}{{\mathcal{O}}}
\newcommand{\Fcal}{{\mathcal{F}}}
\newcommand{\N}{\mathcal{N}}  % for normal density
\newcommand{\bupeta}{\boldsymbol{\upeta}}
\renewcommand{\O}{{\bf O}}
\renewcommand{\P}{{\bf P}}
\newcommand{\Q}{{\bf Q}}
\renewcommand{\S}{{\bf S}}
\newcommand{\Scal}{{\mathcal{S}}}
\newcommand{\T}{{\bf T}}
\newcommand{\Tcal}{{\mathcal{T}}}
\newcommand{\U}{{\bf U}}
\newcommand{\Ucal}{{\mathcal{U}}}
\newcommand{\tUcal}{{\tilde{\Ucal}}}
\newcommand{\V}{{\bf V}}
\newcommand{\W}{{\bf W}}
\newcommand{\Wcal}{{\mathcal{W}}}
\newcommand{\Vcal}{{\mathcal{V}}}
\newcommand{\X}{{\bf X}}
\newcommand{\Xcal}{{\mathcal{X}}}
\newcommand{\Y}{{\bf Y}}
\newcommand{\Ycal}{{\mathcal{Y}}}
\newcommand{\Z}{{\bf Z}}
\newcommand{\Zcal}{{\mathcal{Z}}}

% this is for latex 2.09
% unfortunately, the result is slanted - use Latex2e instead
%\newcommand{\bfLambda}{\mbox{\boldmath$\Lambda$}}
% this is for Latex2e
\newcommand{\bfLambda}{\boldsymbol{\Lambda}}

% Yuan Qi's boldsymbol
\newcommand{\bsigma}{\boldsymbol{\sigma}}
\newcommand{\balpha}{\boldsymbol{\alpha}}
\newcommand{\bpsi}{\boldsymbol{\psi}}
\newcommand{\bphi}{\boldsymbol{\phi}}
\newcommand{\bPhi}{\boldsymbol{\Phi}}
\newcommand{\bbeta}{\boldsymbol{\beta}}
\newcommand{\Beta}{\boldsymbol{\eta}}
\newcommand{\btau}{\boldsymbol{\tau}}
\newcommand{\bvarphi}{\boldsymbol{\varphi}}
\newcommand{\bzeta}{\boldsymbol{\zeta}}

\newcommand{\blambda}{\boldsymbol{\lambda}}
\newcommand{\bLambda}{\mathbf{\Lambda}}

\newcommand{\btheta}{\boldsymbol{\theta}}
\newcommand{\bpi}{\boldsymbol{\pi}}
\newcommand{\bxi}{\boldsymbol{\xi}}
\newcommand{\bSigma}{\boldsymbol{\Sigma}}
\newcommand{\bPi}{\boldsymbol{\Pi}}
\newcommand{\bOmega}{\boldsymbol{\Omega}}

\newcommand{\bx}{{\bf x}}
\newcommand{\bgamma}{\boldsymbol{\gamma}}
\newcommand{\bGamma}{\boldsymbol{\Gamma}}
\newcommand{\bUpsilon}{\boldsymbol{\Upsilon}}

\newcommand{\bmu}{\boldsymbol{\mu}}
\newcommand{\0}{{\bf 0}}

\newcommand{\bs}{\backslash}
\newcommand{\ben}{\begin{enumerate}}
\newcommand{\een}{\end{enumerate}}

 \newcommand{\notS}{{\backslash S}}
 \newcommand{\nots}{{\backslash s}}
 \newcommand{\noti}{{\backslash i}}
 \newcommand{\notj}{{\backslash j}}
 \newcommand{\nott}{\backslash t}
 \newcommand{\notone}{{\backslash 1}}
 \newcommand{\nottp}{\backslash t+1}

\newcommand{\notk}{{^{\backslash k}}}
\newcommand{\notij}{{^{\backslash i,j}}}
\newcommand{\notg}{{^{\backslash g}}}
\newcommand{\wnoti}{{_{\w}^{\backslash i}}}
\newcommand{\wnotg}{{_{\w}^{\backslash g}}}
\newcommand{\vnotij}{{_{\v}^{\backslash i,j}}}
\newcommand{\vnotg}{{_{\v}^{\backslash g}}}
\newcommand{\half}{\frac{1}{2}}
\newcommand{\msgb}{m_{t \leftarrow t+1}}
\newcommand{\msgf}{m_{t \rightarrow t+1}}
\newcommand{\msgfp}{m_{t-1 \rightarrow t}}

\newcommand{\proj}[1]{{\rm proj}\negmedspace\left[#1\right]}

\newcommand{\dif}{\mathrm{d}}
\newcommand{\abs}[1]{\lvert#1\rvert}
\newcommand{\norm}[1]{\lVert#1\rVert}

\newcommand{\mrm}[1]{\mathrm{{#1}}}
\newcommand{\RomanCap}[1]{\MakeUppercase{\romannumeral #1}}
\newcommand{\EE}{\mathbb{E}}
\newcommand{\bbI}{\mathbb{I}}
\newcommand{\bbH}{\mathbb{H}}
\newcommand{\ie}{{\textit{i.e.,}}\xspace}
\newcommand{\eg}{{\textit{e.g.,}}\xspace}
\newcommand{\etc}{{\textit{etc.}}\xspace}
\newcommand{\cmt}[1]{}

\date{September 2026}
\maketitle

\begin{abstract}
Event logs arise in a wide range of real-world processes, capturing not only event activities and timestamps but also multi-modal contextual information. Existing event-sequence models, including many temporal point process approaches, primarily model event activities and timestamps while overlooking heterogeneous context, such as numerical measurements, categorical attributes, textual descriptions, and metadata associated with individual events and entire traces. In this paper, we propose Universal Multi-Modal Traceformer (UMT), a unified framework for incorporating heterogeneous process context into next-event prediction. Built on a Transformer backbone, UMT introduces a universal feature encoder that maps diverse feature types into a shared representation space and handles contextual information at both the event and trace levels. UMT further develops a per-event Perceiver module that dynamically weights contextual features and adaptively integrates them into event-token representations. To accommodate the heavy-tailed and potentially multi-modal distribution of inter-arrival times, UMT represents each interval at multiple temporal scales and jointly predicts the corresponding scale-specific quantities. Experiments on 13 real-world event logs show that UMT improves both next-event activity and time prediction over existing approaches.
\end{abstract}

\section{Introduction}

Event logs arise in a wide range of event-driven applications, including e-commerce platforms, online service workflows,
healthcare portals, IT service desks, and similar web-facing systems.
Each log records timestamped sequences of user and system actions (searches, transactions, support tickets, content edits).
Beyond the activity and timestamp of each event, these logs often contain multi-modal contextual information associated with both individual events and their traces, i.e., event sequences. Such information may include textual descriptions, participant identities or roles, trace-level metadata, numerical measurements, and categorical attributes.

Two research communities have studied this prediction task. In temporal point processes~\citep{daley2003introduction}, conditional intensity functions capture dependencies within and across event activities~\citep{hawkes1971spectra,blundell2012modelling,mei2017neural,zuo2020transformer}, with recent extensions including intensity-free~\citep{Shchur2020IntensityFree} and generative approaches~\citep{ludke2023add,kerrigan2026eventflow}. These models jointly predict event activities and times but typically represent each event by its activity and timestamp alone, with few exceptions that incorporate structured covariates~\citep{meng2024transfeattpp}. In predictive process monitoring~\citep{vanderaalst2016process}, Transformer and recurrent architectures have been applied to activity prefixes of event logs~\citep{bukhsh2021processtransformer,evermann2017predicting,tax2017predictive}, and several methods incorporate structured attributes such as resources~\citep{camargo2019learning}, numerical and categorical event fields~\citep{navarin2017lstm,wang2023mitfm}, or textual descriptions~\citep{pegoraro2021text}. However, these approaches typically concatenate or discretize a log-specific subset of columns, assume a fixed schema, and predict the next activity with a simple scalar duration head. Neither line of work provides a single architecture that (i)~ingests mixed numerical, categorical, and textual fields at both the event and trace level, including missing and previously unseen values, (ii)~weights those fields adaptively for each event, and (iii)~models inter-arrival times whose mass may span minutes to weeks. As a result, deploying a next-event predictor on a new log still requires log-specific feature selection and preprocessing, limiting the practical applicability of existing methods.

To address this limitation, we propose the Universal Multi-Modal Traceformer (UMT), a framework that incorporates heterogeneous contextual information into joint next-event activity and inter-arrival time prediction, and can be applied to new event logs without log-specific feature engineering. Our main contributions are threefold.

First, we introduce a universal feature encoder that maps heterogeneous inputs, including text, categorical attributes, and numerical measurements, into a shared latent space. Modality-specific encoding paths handle the characteristics of each feature type while supporting missing or previously unseen values. The encoder processes fields at both the event and trace level, so the same architecture applies regardless of which columns a particular log contains.

Second, we develop a per-event Perceiver module for dynamic feature weighting and integration. For each event, an embedding of its activity and observed inter-arrival time serves as a query that cross-attends to the associated contextual features. This mechanism adaptively weights features according to their relevance, allowing the model to exploit informative fields and ignore irrelevant or absent ones without manual column selection. A causal Transformer then models the resulting event sequence to predict the activity and time of the next event.

Third, we introduce a multi-scale representation and prediction scheme for inter-arrival times to better accommodate their heavy-tailed and potentially multi-modal distributions. Each interval is decomposed across temporal scales, such as days, hours, and minutes, and the model jointly predicts the corresponding scale-specific quantities. Cyclic components, such as hours and minutes, are represented as angles and trained with a cosine-based loss that avoids discontinuities at temporal boundaries.

We evaluate UMT on $13$ real-world event logs spanning business processes, IT incidents, energy consumption, healthcare, food safety, and transportation. UMT achieves the highest next-activity accuracy on $11$ of $13$ datasets and the lowest duration MAE on $9$ of $13$ (Tables~\ref{tab:accuracy}--\ref{tab:mae}), with MAE reductions of $27\%$--$40\%$ over the strongest baseline on several metadata-rich logs. Ablations confirm that each component contributes: the multi-scale time head reduces MAE by $12\%$ on average over a log-normal baseline (Table~\ref{tab:abl-timehead}), and the feature encoder adds up to $18$ accuracy points on logs with rich contextual fields (Table~\ref{tab:abl-features}). Perceiver-based fusion proves robust to irrelevant context: when padded to $100$ columns, it loses at most $3.6$ accuracy points, compared with $13.5$ for additive fusion (Table~\ref{tab:abl-scaling}).

\section{Related Work}

Temporal point processes~\citep{daley2003introduction,daley2008introduction} are the dominant framework for event modeling. Early approaches often rely on Poisson processes~\citep{lawless1987regression,grandell2006doubly,charlin2015dynamic,gopalan2014content,gopalan2015scalable}, which assume independent increments and therefore do not model interactions among events in a trace. Hawkes processes~\citep{hawkes1971spectra} relax this assumption by capturing excitation effects between events, leading to numerous extensions and applications in influence modeling, scalability, and structured event prediction~\citep{blundell2012modelling,wang2017predicting,yang2017decoupling,xu2018benefits,zhe2018stochastic,pan2020scalable,pan2021self}.

More recently, neural point processes have used deep architectures to parameterize conditional intensity functions. Representative models encode event histories with recurrent neural networks or continuous-time recurrent states, such as RMTPP~\citep{du2016recurrent} and Neural Hawkes Processes~\citep{mei2017neural}, or model cumulative intensities with monotonic networks to avoid explicit numerical integration~\citep{omi2019fully,sill1997monotonic}. Attention-based variants, including Transformer Hawkes Processes~\citep{zuo2020transformer}, Self-Attentive Hawkes Processes~\citep{zhang2020self}, and Attentive Neural Hawkes Processes~\citep{yang2022transformer}, represent events and their types as tokens and use causal attention to encode historical dependencies. Other recent work combines structured and neural point-process components~\citep{yuan2025residual}, while EasyTPP provides an open-source benchmark for neural point-process models~\citep{xueeasytpp}.
Another line of work develops intensity-free generative models for temporal event generation, including approaches based on normalizing flows~\citep{Shchur2020IntensityFree}, diffusion models~\citep{lin2022exploring,ludke2023add}, and flow matching~\citep{kerrigan2026eventflow,ludke2026editbased}.

Process mining analyzes event logs to discover, monitor, and improve operational processes~\citep{vanderaalst2016process}. Within this field, predictive process monitoring uses prefixes of ongoing traces to forecast future activities, timestamps, remaining times, and process outcomes. Early deep-learning approaches use LSTMs to encode activity sequences~\citep{evermann2017predicting,tax2017predictive}, with later work adopting Transformer backbones~\citep{bukhsh2021processtransformer}. Several methods go beyond activity and time by incorporating additional event attributes: Navarin et al.\ use numerical features for remaining-time prediction~\citep{navarin2017lstm}; Camargo et al.\ embed resource roles alongside activities~\citep{camargo2019learning}; MiTFM discretizes all event-log columns and fuses them via multi-head attention~\citep{wang2023mitfm}; and Pegoraro et al.\ encode textual event descriptions with a language model~\citep{pegoraro2021text}. In the TPP literature, TransFeat-TPP incorporates structured covariates into a Transformer point process and learns feature-importance rankings~\citep{meng2024transfeattpp}. However, these methods are generally designed for particular feature types or prediction tasks: each targets a specific combination of input modalities, assumes a fixed schema, or handles only part of the prediction problem (e.g., next activity without time, or remaining time without next activity).

UMT differs in three respects: it maps mixed numerical, categorical, and textual fields into a shared token space without requiring a fixed schema; it uses a Perceiver module whose event-conditioned cross-attention adapts to whichever fields are present; and it jointly predicts next-event activity and inter-arrival time through a multi-scale time head designed for heavy-tailed durations.

\begin{figure*}
    \centering
    \includegraphics[width=\textwidth,trim={0.9cm 0 0.9cm 0},clip]{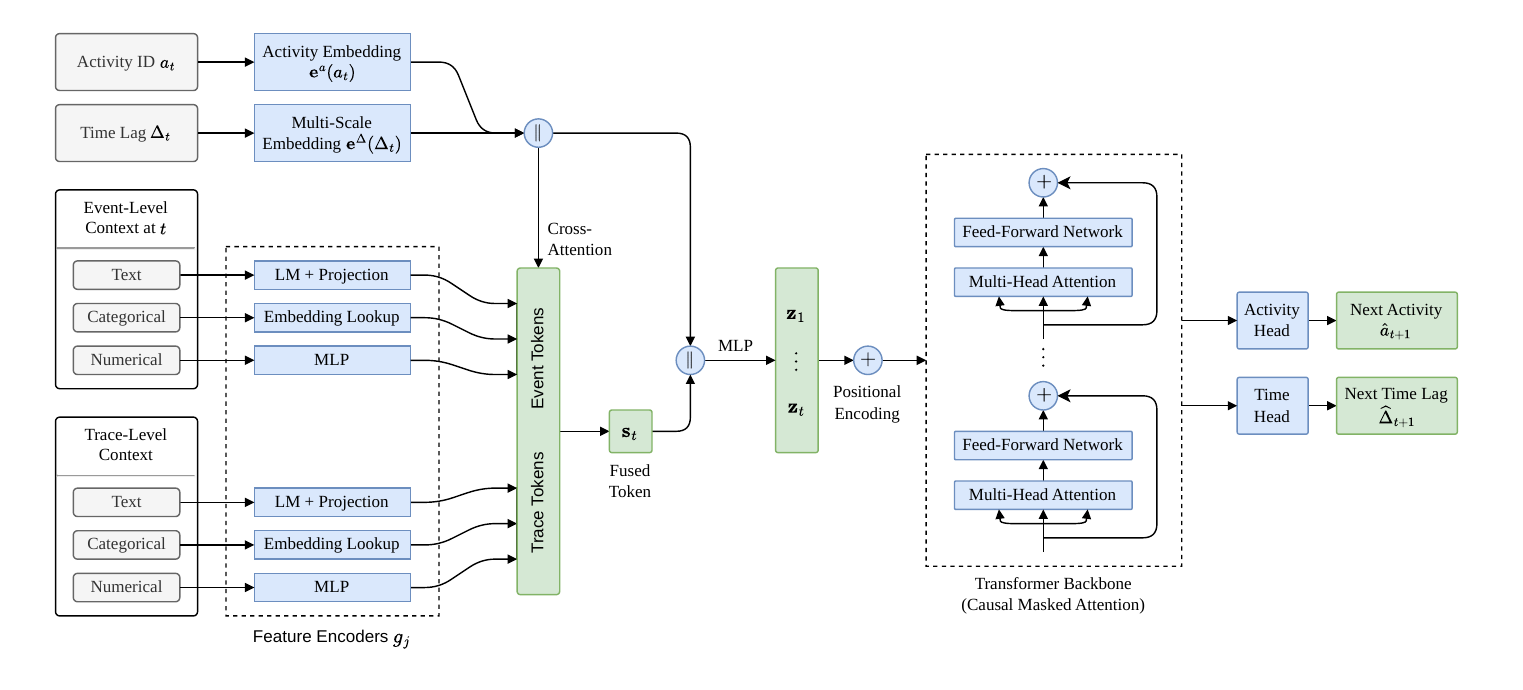}
    \caption{Architecture overview. Universal feature encoders map trace-level
    and event-level fields to individual tokens. For each event, an
    activity--time query cross-attends to the available feature tokens in a
    Perceiver module, producing a context-aware event token. A causal
    Transformer then processes the resulting sequence and feeds separate
    next-event-activity and multi-scale inter-arrival-time prediction heads.}
    \Description{A left-to-right architecture diagram. Trace-level and
    event-level contextual fields are independently encoded into feature
    tokens. Per-event activity and time embeddings form a query that
    cross-attends to those tokens in a Perceiver fusion module. The resulting
    event-token sequence enters a causal Transformer and branches to activity
    and multi-scale time prediction heads.}
    \label{fig:architecture}
\end{figure*}

\section{Methodology}
\label{sec:method}

\subsection{Problem Formulation and Model Overview}

Let an event log contain a collection of traces. We represent each trace as
\begin{equation}
\Gamma =
\left(
\mathbf{c},
\bigl((a_t,\Delta_t,\mathbf{x}_t)\bigr)_{t=1}^{T}
\right),
\end{equation}
where $\mathbf{c}=(c_1,\ldots,c_J)$ denotes trace-level contextual fields, $a_t \in \mathcal{A}$ is the activity of the $t$-th event, $\Delta_t \geq 0$ is its inter-arrival time from the preceding event, and $\mathbf{x}_t=(x_{t,1},\ldots,x_{t,K_t})$ collects the event-level contextual fields. Both $\mathbf{c}$ and $\mathbf{x}_t$ may include heterogeneous fields, such as numerical measurements, categorical attributes, and textual descriptions. For simplicity, we set $\Delta_1 = 0$.

Given a trace prefix $\Gamma_{\leq t}$, our goal is to predict the activity $a_{t+1}$ and inter-arrival time $\Delta_{t+1}$ of the next event. The proposed Universal Multi-Modal Traceformer (UMT) consists of four main components. A universal feature encoder first maps heterogeneous contextual fields into a shared latent space. A per-event Perceiver module then dynamically weights and integrates the available context into each event representation. A causal Transformer captures temporal dependencies across the resulting event sequence. Finally, separate prediction heads estimate the next-event activity and its inter-arrival time, with the latter represented through a multi-scale decomposition.

\subsection{Universal Multi-Modal Feature Encoder}
\label{sec:feature-encoder}

At event position $t$, we collect the event-level fields in $\mathbf{x}_t$ together with the trace-level fields in $\mathbf{c}$. Let $u_{t,j}$ range over the combined set $\{c_1,\ldots,c_J,\, x_{t,1},\ldots,x_{t,K_t}\}$. UMT applies a feature-specific encoder $g_j$ and maps each field to the common model dimension $d$:
\begin{equation}
\mathbf{f}_{t,j}=g_j(u_{t,j})\in\mathbb{R}^{d}.
\label{eq:feature-encoding}
\end{equation}

The form of $g_j$ depends on the modality of the field. A categorical value is mapped through a learnable embedding table containing dedicated entries for missing and previously unseen categories. A numerical value is represented jointly with its missing-value indicator and transformed by a small multilayer perceptron. Textual content is encoded using a lightweight pretrained language model, pooled into a fixed-dimensional representation, and projected to the model dimension. Additional modalities can be incorporated by introducing corresponding encoding paths without changing the subsequent architecture.

The encoded fields are retained as individual feature tokens:
\begin{equation}
\mathbf{F}_t=
\left[
\mathbf{f}_{t,1};
\ldots;
\mathbf{f}_{t,N_t}
\right]
\in\mathbb{R}^{N_t\times d},
\label{eq:feature-matrix}
\end{equation}
where $N_t = J + K_t$ is the total number of fields at event $t$. Each trace-level field is encoded once and its token appears identically in $\mathbf{F}_t$ for all $t$, allowing the downstream Perceiver to weight trace-level attributes differently at each event rather than committing to a single static summary.
Mapping all modalities to a shared space provides a uniform interface for downstream fusion. Preserving each field as a separate token avoids premature aggregation and retains feature-specific information for adaptive fusion.

\subsection{Per-Event Perceiver Fusion}
\label{sec:perceiver}

The importance of a contextual field may vary across events and traces. For example, a customer comment may be particularly informative for one activity, whereas a numerical measurement may be more useful for another. We therefore develop a per-event Perceiver~\cite{Jaegle21} module that performs event-conditioned soft weighting and integration of contextual features.

For each observed event, we construct an event-specific query based solely on its activity and inter-arrival time:
\begin{equation}
\mathbf{q}_t =
\mathbf{W}_q
\left[
\mathbf{e}^{a}(a_t);
\mathbf{e}^{\Delta}(\Delta_t)
\right]
+\mathbf{b}_q \in \mathbb{R}^{d_q},
\label{eq:event-query}
\end{equation}
where $\mathbf{e}^{a}$ is a learnable activity embedding and $\mathbf{e}^{\Delta}$ encodes the observed interval $\Delta_t$ using the multi-scale representation introduced later. 

The event query cross-attends to all contextual feature tokens: $\mathbf{K}_t=\mathbf{F}_t\mathbf{W}_k$, $\mathbf{V}_t=\mathbf{F}_t\mathbf{W}_v$, and
\begin{align}
\boldsymbol{\alpha}_t=
\operatorname{softmax}
\left(
\frac{\mathbf{K}_t\mathbf{q}_t}
{\sqrt{d_q}}
\right)
\in\mathbb{R}^{N_t},\quad 
\mathbf{s}_t=\boldsymbol{\alpha}_t^\top\mathbf{V}_t.
\end{align}
The cross-attention weights ${\boldsymbol \alpha}_t$ provide a soft, event-dependent measure of feature importance. The model can thus adaptively integrate different subsets of contextual information. We then combine the fused context with the original activity and inter-arrival-time embeddings to form the event token: 
\begin{align}
    \mathbf{z}_t = \mathbf{W}_z \left[\mathbf{s}_t;
\mathbf{e}^{a}(a_t);
\mathbf{e}^{\Delta}(\Delta_t)
\right] + \mathbf{b}_z.
\end{align}

This design has two additional advantages. First, it supports \textit{varying} numbers and types of contextual fields across datasets. Second, a single event query attends to the feature tokens, so the attention computation grows linearly rather than quadratically in the number of contextual features. The resulting $\mathbf{z}_t$ serves as a \textit{context-aware} representation of event $t$.

\subsection{Causal Trace Modeling}

After contextual fusion, positional embeddings are added to the event representations, and the sequence is processed by an $L$-layer causal Transformer:
\begin{equation}
(\mathbf{h}_1,\ldots,\mathbf{h}_T)
= 
\operatorname{Transformer}_{\mathrm{causal}}
(\mathbf{z}_1+\mathbf{p}_1,\ldots,
\mathbf{z}_T+\mathbf{p}_T), \notag 
\end{equation}
where causal masking ensures that $\mathbf{h}_t$ depends only on the prefix through event $t$. Thus, $\mathbf{h}_t$ summarizes both the historical event sequence and the heterogeneous context relevant to predicting the next event.

The next-event activity distribution is produced by
\begin{equation}
p(a_{t+1}\mid\Gamma_{\leq t})
=
\operatorname{softmax}
\left(
\mathbf{W}_a\mathbf{h}_t+\mathbf{b}_a
\right).
\label{eq:activity-head}
\end{equation}
The causal architecture allows predictions for all trace prefixes to be trained in parallel while preventing information leakage from future events.

\subsection{Multi-Scale Inter-Arrival Time Modeling}
\label{sec:multiscale-time}

Inter-arrival times in real-world processes are often highly skewed and may span several orders of magnitude, from minutes to multiple days or weeks. A single regression or distributional head must then capture both frequent short intervals and rare long delays. This is difficult for both statistical modeling and numerical optimization.

UMT addresses these challenges by decomposing each interval into components with different statistical structures. Let
$S_d=86{,}400$, $S_h=3{,}600$, and $S_m=60$ denote the numbers of seconds in a day, hour, and minute. For each inter-arrival time $\Delta_t$ (measured in seconds), we define
\begin{align}
D_t =
\left\lfloor\frac{\Delta_t}{S_d}\right\rfloor, \quad
H_t =
\left\lfloor
\frac{\Delta_t-D_t S_d}{S_h}
\right\rfloor, \quad
M_t =
\frac{\Delta_t-D_t S_d-H_t S_h}{S_m}. \notag
\label{eq:time-decomposition}
\end{align}
Thus, $D_t\in\mathbb{Z}_{\geq 0}$, $H_t\in\{0,\ldots,23\}$, and $M_t\in[0,60)$, with
\begin{equation}
\Delta_t=D_t \cdot S_d+H_t \cdot S_h+M_t \cdot S_m.
\label{eq:time-reconstruction}
\end{equation}

This decomposition separates the unbounded and heavy-tailed day count from the bounded within-day components. It therefore allows each prediction head to focus on a substantially narrower and more appropriate target space.

\paragraph{Day-scale prediction.}
The day component is a nonnegative and potentially overdispersed count. From $\mathbf{h}_t$, a day-scale head $f_d$ produces two unconstrained outputs. These are transformed elementwise into the positive mean $\mu_t$ and dispersion $r_t$ of a Negative Binomial (NB) distribution:
\begin{equation}
(\mu_t,r_t)
=
\operatorname{softplus}
\left(
f_d(\mathbf{h}_t)
\right).
\end{equation}
Under the mean--dispersion parameterization, we model
\begin{equation}
D_{t+1}\sim
\operatorname{NB}(\mu_t,r_t),
\label{eq:day-distribution}
\end{equation}
whose variance is $\mu_t+\mu_t^2/r_t$. Unlike a Poisson model, this parameterization allows the variance to exceed the mean. It is therefore better suited to overdispersed and long-tailed delays while still providing a full predictive distribution.

\paragraph{Cyclic hour- and minute-scale prediction.}
The residual hour and minute components are bounded and periodic. Treating them as ordinary scalar targets introduces artificial discontinuities at their wraparound boundaries; for example, minute values near $59$ and $0$ are numerically far apart despite being adjacent on the underlying cycle. We therefore represent both components as unit vectors. Their target representations are
\begin{align}
\mathbf{u}_{h,t+1}
&=
\left[
\cos\left(\frac{2\pi H_{t+1}}{24}\right),
\sin\left(\frac{2\pi H_{t+1}}{24}\right)
\right], \notag \\
\mathbf{u}_{m,t+1}
&=
\left[
\cos\left(\frac{2\pi M_{t+1}}{60}\right),
\sin\left(\frac{2\pi M_{t+1}}{60}\right)
\right].
\label{eq:cyclic-targets}
\end{align}
Two prediction heads, $f_h$ and $f_m$, map the contextual representation $\mathbf{h}_t$ to normalized two-dimensional vectors:
\begin{equation}
\widehat{\mathbf{u}}_{s,t+1} =
\frac{f_s(\mathbf{h}_t)}
{\lVert f_s(\mathbf{h}_t)\rVert_2},
\qquad s\in \{h,m\}.
\label{eq:cyclic-heads}
\end{equation}
This circular representation preserves proximity across scale boundaries when combined with the corresponding higher-scale component. It also enables a bounded and smooth cosine-based objective. Direct scalar regression near wraparound boundaries can produce disproportionately large errors; the circular formulation avoids this.

\subsection{Learning and Prediction}

For each prefix ending at event $t$, the activity loss is
\begin{equation}
\mathcal{L}_{a,t}
=
-\log p(a_{t+1}\mid\Gamma_{\leq t}).
\end{equation}
The multi-scale time loss combines the Negative Binomial negative log-likelihood with two cosine-distance terms:
\begin{align}
\mathcal{L}_{\Delta,t}
=&
-\log p_{\mathrm{NB}}
\left(
D_{t+1}\mid\mu_t,r_t
\right) \nonumber \\
&+
\lambda_h
\left(
1-
\widehat{\mathbf{u}}_{h,t+1}^{\top}
\mathbf{u}_{h,t+1}
\right) \nonumber \\
&+
\lambda_m
\left(
1-
\widehat{\mathbf{u}}_{m,t+1}^{\top}
\mathbf{u}_{m,t+1}
\right).
\label{eq:time-loss}
\end{align}

For a trace of length $T$, the complete training objective, averaged over its $T-1$ valid prediction positions, is
\begin{equation}
\mathcal{L} =
\frac{1}{T-1}
\sum_{t=1}^{T-1}
\left(
\mathcal{L}_{a,t}
+
\lambda_{\Delta}\mathcal{L}_{\Delta,t}
\right).
\label{eq:joint-objective}
\end{equation}
During mini-batch training, we compute this objective for each trace and average it across all traces in the mini-batch.

At inference, the next activity is selected from Equation~\eqref{eq:activity-head}. For time prediction, the day component is summarized by the mean of the Negative Binomial distribution. The hour and minute components are recovered from the predicted angles:
\begin{align}
\widehat{H}_{t+1}
&=
\frac{24}{2\pi}
\left[
\operatorname{atan2}
\!\left(
(\widehat{\mathbf{u}}_{h,t+1})_2,\,
(\widehat{\mathbf{u}}_{h,t+1})_1
\right)
\right]_{2\pi},\notag \\ 
\widehat{M}_{t+1}
&=
\frac{60}{2\pi}
\left[
\operatorname{atan2}
\!\left(
(\widehat{\mathbf{u}}_{m,t+1})_2,\,
(\widehat{\mathbf{u}}_{m,t+1})_1
\right)
\right]_{2\pi},
\end{align}
where $[\cdot]_{2\pi}$ maps an angle to $[0,2\pi)$. The final interval prediction is reconstructed as
\begin{equation}
\widehat{\Delta}_{t+1}
=
\widehat{D}_{t+1}\cdot S_d +
\widehat{H}_{t+1}\cdot S_h +
\widehat{M}_{t+1}\cdot S_m.
\end{equation}
Together, these four components let UMT incorporate heterogeneous context from event logs with different modalities and feature sets.

\begin{table}[t]
\centering
\caption{Dataset statistics. All datasets are publicly available.}
\medskip
\label{tab:datasets}
\footnotesize
\setlength{\tabcolsep}{3pt}
\begin{tabular}{@{}lrrrrrr@{}}
\toprule
\textbf{Dataset} & \makecell{\textbf{Num.}  \\ \textbf{cases}} & \makecell{\textbf{Num.}  \\ \textbf{events}} & \makecell{\textbf{Num.}  \\ \textbf{act.}} & \makecell{\textbf{Mean} \\ \textbf{ev./case}} & \makecell{\textbf{Max} \\ \textbf{ev./case}} & \makecell{\textbf{Mean} \\ \textbf{dur.\ (d)}} \\
\midrule
BPI-2012       & 13087  & 262200  & 24  & 24  & 175  & 8.6  \\
BPI-2013       & 7553   & 65532   & 4   & 4   & 123  & 12.1 \\
BPI-2015-1     & 1199   & 52217   & 398 & 44  & 101  & 95.7  \\
BPI-2015-2     & 830    & 44352   & 410 & 53  & 132  & 160.5 \\
BPI-2017-Off   & 42995  & 193849  & 8   & 5   & 5    & 19.1  \\
BPI-2020-Dom   & 10366  & 56303   & 17  & 5   & 24   & 11.7  \\
BPI-2020-Int   & 6449   & 72151   & 34  & 11  & 27   & 86.5  \\
BPI-2020-Req   & 6814   & 36724   & 19  & 5   & 20   & 12.1  \\
National-Grid  & 1827   & 43829   & 5   & 24  & 26   & 1.0   \\
IT-Incidents   & 3317   & 16347   & 13  & 5   & 20   & 32.2  \\
Medical-Appts  & 6830   & 19699   & 4   & 3   & 9    & 16.8  \\
Chicago-Food   & 44167  & 264066  & 99  & 6   & 682  & 1286.4 \\
Train-Delays   & 54423  & 406113  & 3   & 7   & 24   & 0.04  \\
\bottomrule
\end{tabular}
\end{table}

\begin{table*}
\caption{Next-activity prediction accuracy ($\uparrow$) on $13$ real-world event logs. Bold indicates the best result per dataset. $(\dagger)$ denotes feature-stripped variants that receive only activity and time embeddings. Results are mean $\pm$ standard deviation over $5$ seeds.}
\medskip
\label{tab:accuracy}
\centering
\footnotesize
\setlength{\tabcolsep}{2.5pt}
\begin{tabular}{l r@{\,${}\pm{}$\,}l r@{\,${}\pm{}$\,}l r@{\,${}\pm{}$\,}l r@{\,${}\pm{}$\,}l r@{\,${}\pm{}$\,}l r@{\,${}\pm{}$\,}l r@{\,${}\pm{}$\,}l}
\toprule
   & \multicolumn{2}{c}{UMT} & \multicolumn{2}{c}{Neural H.} & \multicolumn{2}{c}{Trans. H.} & \multicolumn{2}{c}{IFTPP} & \multicolumn{2}{c}{GRU} & \multicolumn{2}{c}{UMT\,$(\dagger)$} & \multicolumn{2}{c}{GRU\,$(\dagger)$} \\
\midrule
  BPI-2012 & \textbf{0.86} & \textbf{0.00} & 0.69 & 0.01 & 0.18 & 0.13 & 0.82 & 0.01 & 0.03 & 0.02 & 0.84 & 0.00 & 0.02 & 0.01 \\
  BPI-2013 & \textbf{0.76} & \textbf{0.00} & 0.55 & 0.00 & 0.42 & 0.04 & 0.60 & 0.00 & 0.37 & 0.11 & 0.60 & 0.01 & 0.18 & 0.14 \\
  BPI-2015-1 & \textbf{0.53} & \textbf{0.01} & 0.13 & 0.06 & 0.03 & 0.01 & 0.31 & 0.01 & 0.00 & 0.01 & \textbf{0.53} & \textbf{0.01} & 0.00 & 0.01 \\
  BPI-2015-2 & 0.49 & 0.02 & 0.12 & 0.03 & 0.00 & 0.00 & 0.37 & 0.02 & 0.01 & 0.01 & \textbf{0.54} & \textbf{0.01} & 0.00 & 0.00 \\
  BPI-2017-Off & \textbf{0.92} & \textbf{0.00} & 0.83 & 0.00 & 0.82 & 0.01 & 0.74 & 0.00 & \textbf{0.92} & \textbf{0.00} & 0.75 & 0.00 & 0.75 & 0.00 \\
  BPI-2020-Dom & \textbf{0.89} & \textbf{0.00} & 0.86 & 0.00 & 0.43 & 0.15 & \textbf{0.89} & \textbf{0.00} & 0.08 & 0.11 & \textbf{0.89} & \textbf{0.00} & 0.15 & 0.23 \\
  BPI-2020-Int & \textbf{0.89} & \textbf{0.00} & 0.86 & 0.00 & 0.24 & 0.13 & 0.88 & 0.00 & 0.06 & 0.05 & 0.88 & 0.00 & 0.02 & 0.05 \\
  BPI-2020-Req & \textbf{0.90} & \textbf{0.00} & 0.86 & 0.00 & 0.35 & 0.21 & 0.88 & 0.00 & 0.04 & 0.05 & 0.88 & 0.00 & 0.11 & 0.14 \\
  National-Grid & \textbf{0.86} & \textbf{0.03} & 0.80 & 0.00 & 0.79 & 0.01 & 0.74 & 0.02 & 0.20 & 0.10 & 0.81 & 0.00 & 0.21 & 0.12 \\
  IT-Incidents & \textbf{0.71} & \textbf{0.01} & 0.59 & 0.01 & 0.34 & 0.13 & 0.66 & 0.00 & 0.16 & 0.12 & 0.69 & 0.00 & 0.12 & 0.10 \\
  Medical-Appts & \textbf{1.00} & \textbf{0.00} & 0.98 & 0.00 & 0.66 & 0.23 & 0.94 & 0.00 & 0.31 & 0.14 & 0.94 & 0.00 & 0.23 & 0.14 \\
  Chicago-Food & \textbf{0.76} & \textbf{0.00} & 0.72 & 0.00 & 0.41 & 0.00 & 0.42 & 0.00 & 0.00 & 0.01 & 0.42 & 0.00 & 0.01 & 0.02 \\
  Train-Delays & \textbf{0.96} & \textbf{0.00} & 0.91 & 0.00 & 0.89 & 0.00 & 0.91 & 0.00 & 0.41 & 0.45 & 0.91 & 0.00 & 0.08 & 0.04 \\
\bottomrule
\end{tabular}
\end{table*}

\section{Experiments}
\label{sec:experiments}

We evaluate our model on the standard next-event prediction task,
i.e.\ given a prefix of events $(a_1, t_1), \ldots, (a_t, t_t)$ in a
running case, jointly predict the activity label $a_{t+1}$ and the
duration $\Delta t_{t+1} = t_{t+1} - t_t$ until the next event. We
first compare against established neural point-process baselines on a
diverse benchmark of $13$ event logs
(Section~\ref{sec:main-results}), then isolate the contribution of
each architectural component through targeted ablations
(Section~\ref{sec:ablation}).

\subsection{Experimental Setup}
\label{sec:setup}

\paragraph{Datasets}
We evaluate on $13$ publicly available event logs from several domains with different temporal scales (Table~\ref{tab:datasets}). Eight logs come from the Business Process Intelligence (BPI) Challenges~\citep{bpichallenge}: loan applications (BPI-2012, BPI-2017-Off), IT incident management (BPI-2013), building permits (BPI-2015-1, BPI-2015-2), and travel and reimbursement workflows (BPI-2020-Dom, BPI-2020-Int, BPI-2020-Req). For BPI-2015 we adopt the coarsened activity vocabulary that groups fine-grained administrative steps into high-level phases; for BPI-2020-Req we use the publicly recommended preprocessing that removes auxiliary labels.

Beyond these process-mining benchmarks, we include five additional public event logs:
\emph{National-Grid}~\citep{dataset_nationalgrid}, hourly electricity consumption from a national grid discretized into consumption-level activities;
\emph{IT-Incidents}~\citep{dataset_itincidents}, a university IT service desk log;
\emph{Medical-Appts}~\citep{dataset_medicalappointments}, a patient appointment lifecycle log including no-shows;
\emph{Chicago-Food}~\citep{dataset_chicagofood}, food-establishment inspection histories from the Chicago Department of Public Health; and
\emph{Train-Delays}~\citep{dataset_traindelays}, NJ Transit and Amtrak rail operations with per-event delay measurements.
Full dataset statistics are given in Table~\ref{tab:datasets}.

For every dataset we sort cases by their start timestamp and reserve
the most recent $15\%$ as the held-out test set. The remaining $85\%$
of cases is further split $85$/$15$ (uniformly at random, seed~$42$)
into training and validation folds. Splitting by case avoids leakage
between training and test events that would otherwise occur with a
naive event-level split. The temporal ordering preserves the
realistic deployment setting where past cases are used to predict
future ones. Sequences longer than the $95$-th percentile of training
case lengths are left-truncated, with an absolute upper cap of $150$
events.

\paragraph{Evaluation metrics}
We report standard process-prediction metrics:
\emph{accuracy} for next-activity prediction, and \emph{mean absolute error} (MAE) for the predicted
inter-event time $\widehat{\Delta t}$ versus the ground truth
$\Delta t$. Both metrics are computed over all non-padded positions
except the first position of each trace (which has no preceding context). The unit of MAE is chosen per dataset according to its
characteristic time scale and is indicated in parentheses:
(h)~=~hours, (d)~=~days, (min)~=~minutes.
An entry of $\infty$ indicates that the model produced numerically
divergent predictions (e.g., NaN or infinite values), rendering
the MAE undefined.
IFTPP produced divergent duration predictions on $11$ of $13$
datasets; the sole exception is BPI-2013
($87.01 \!\pm\! 140.15$~days). Because the column would consist
almost entirely of $\infty$ entries, we omit IFTPP from
Table~\ref{tab:mae}.
We repeat all experiments 5 times with different random seeds and
report the mean and standard deviation.
MAE is computed from the mean of the predicted inter-arrival time distribution: the Negative Binomial mean $\mu_t$ for the day component in our multi-scale head and the log-normal median for the baseline head. The best validation epoch is
selected by the joint \emph{rank-sum} of validation accuracy and
validation MAE (in the respective dataset unit). This avoids biasing model selection towards either head.

\begin{table*}
\caption{Inter-arrival time MAE ($\downarrow$) on $13$ real-world event logs. Unit column: h = hours, d = days, min = minutes. Bold indicates the best result per dataset. $(\dagger)$ denotes feature-stripped variants. IFTPP is omitted because it produced numerically divergent time predictions on $11$ of $13$ datasets. Results are mean $\pm$ standard deviation over $5$ seeds.}
\medskip
\label{tab:mae}
\centering
\footnotesize
\setlength{\tabcolsep}{0.9pt}
\begin{tabular}{lc r@{\,${}\pm{}$\,}l r@{\,${}\pm{}$\,}l r@{\,${}\pm{}$\,}l r@{\,${}\pm{}$\,}l r@{\,${}\pm{}$\,}l r@{\,${}\pm{}$\,}l}
\toprule
   & Unit & \multicolumn{2}{c}{UMT} & \multicolumn{2}{c}{Neural H.} & \multicolumn{2}{c}{Trans. H.} & \multicolumn{2}{c}{GRU} & \multicolumn{2}{c}{UMT\,$(\dagger)$} & \multicolumn{2}{c}{GRU\,$(\dagger)$} \\
\midrule
  BPI-2012 & h & 7.56 & 0.07 & \textbf{7.23} & \textbf{0.24} & 13.24 & 2.38 & 9.69 & 0.03 & 8.26 & 0.07 & 9.68 & 0.00 \\
  BPI-2013 & d & \textbf{0.57} & \textbf{0.04} & 0.89 & 0.04 & 1.75 & 0.29 & 1.11 & 0.00 & 0.84 & 0.01 & 1.11 & 0.00 \\
  BPI-2015-1 & h & 26.55 & 2.45 & \textbf{22.47} & \textbf{2.47} & 98.20 & 122.82 & 32.71 & 0.00 & 27.64 & 1.02 & 32.71 & 0.00 \\
  BPI-2015-2 & h & 28.76 & 3.23 & \textbf{25.45} & \textbf{1.49} & \multicolumn{2}{c}{$\infty$} & 38.83 & 0.00 & 28.17 & 1.62 & 38.83 & 0.00 \\
  BPI-2017-Off & d & \textbf{3.76} & \textbf{0.07} & 5.51 & 0.09 & 6.04 & 0.21 & 4.03 & 0.13 & 4.47 & 0.02 & 4.44 & 0.01 \\
  BPI-2020-Dom & h & \textbf{53.81} & \textbf{0.26} & 55.31 & 0.59 & 55.49 & 5.61 & 66.24 & 3.57 & 54.53 & 0.31 & 68.96 & 1.00 \\
  BPI-2020-Int & d & 4.42 & 0.18 & 5.86 & 0.23 & 6.38 & 0.92 & 4.65 & 0.15 & \textbf{4.28} & \textbf{0.13} & 4.71 & 0.10 \\
  BPI-2020-Req & h & \textbf{65.09} & \textbf{0.31} & 66.57 & 0.82 & 68.75 & 7.00 & 80.21 & 6.19 & 67.30 & 1.83 & 84.62 & 1.81 \\
  National-Grid & h & 0.54 & 0.12 & 92.12 & 136.08 & 2.61 & 3.24 & \textbf{0.02} & \textbf{0.01} & 0.48 & 0.08 & 0.03 & 0.02 \\
  IT-Incidents & d & \textbf{5.04} & \textbf{0.06} & 6.56 & 0.59 & 12.70 & 3.56 & 5.26 & 0.02 & 5.24 & 0.00 & 5.26 & 0.01 \\
  Medical-Appts & d & \textbf{89.92} & \textbf{0.09} & 105.94 & 1.50 & 135.21 & 40.98 & 141.94 & 0.05 & 96.41 & 0.15 & 142.11 & 0.28 \\
  Chicago-Food & d & \textbf{81.50} & \textbf{1.35} & 111.25 & 7.66 & 120.27 & 18.37 & 136.37 & 5.35 & 121.34 & 1.47 & 131.89 & 3.74 \\
  Train-Delays & min & \textbf{3.97} & \textbf{0.04} & 5.36 & 0.04 & 5.01 & 0.54 & 5.96 & 0.94 & 4.59 & 0.01 & 4.97 & 0.23 \\
\bottomrule
\end{tabular}
\end{table*}

\subsection{Methods}
\label{sec:baselines}

\paragraph{UMT (ours)}
We train UMT with $d_{\text{model}}=64$, a $4$-layer, $4$-head
pre-norm causal Transformer with feed-forward width $64$ and dropout $0.1$.
Perceiver fusion uses $4$-head cross-attention with feed-forward width $256$.
Textual fields are embedded by a frozen \texttt{distilbert-base-uncased}
(mean-pooled, $768$-d) and projected to $d_{\text{model}}$.
Training uses AdamW at a constant learning rate of $10^{-3}$, weight
decay $0.03$, batch size $256$, gradient-norm clipping at $1.0$,
time-loss weight $\lambda_{\Delta}=0.1$, and early stopping with
patience $50$ (max $1000$ epochs).
All models are trained on a single NVIDIA L40S GPU.

\paragraph{Baselines}
We compare against four neural baselines that span the dominant
paradigms for next-event prediction in temporal point processes:
\begin{itemize}
    \item \textbf{Neural Hawkes}~\citep{mei2017neural}: A continuous-time
    LSTM that models mutually exciting event dynamics, using a neurally
    parameterized intensity function with exponential memory kernels.
    \item \textbf{Transformer Hawkes}~\citep{zuo2020transformer}: A
    self-attention-based point process that replaces the recurrent
    backbone of Neural Hawkes with a Transformer encoder and temporal
    positional encodings.
    \item \textbf{IFTPP}~\citep{Shchur2020IntensityFree}: An intensity-free
    model that directly parameterizes the conditional inter-event time
    distribution using normalizing flows, bypassing intensity-function
    integration.
    \item \textbf{GRU}~\citep{du2016recurrent}: A GRU-based recurrent
    marked temporal point process that embeds event history into a
    fixed-size vector for next-event prediction.
\end{itemize}
All baselines are trained on the identical splits and evaluation
protocol described above.
We additionally include \emph{feature-stripped}
variants of our own model and the GRU baseline
(denoted~$\dagger$ in Tables~\ref{tab:accuracy}--\ref{tab:mae}).
These variants receive only the activity token and time embedding as input,
isolating the effect of our multimodal feature encoder from the
remaining architecture.

\subsection{Main Results}
\label{sec:main-results}

Table~\ref{tab:accuracy} reports next-activity accuracy and
Table~\ref{tab:mae} reports duration MAE for all methods.
The bold entry per row marks the best result on that dataset.

\paragraph{Activity prediction}
UMT achieves the highest next-activity accuracy on $11$ of $13$
datasets, often by a large margin. On BPI-2013, National-Grid,
and Medical-Appts it exceeds the strongest baseline by $7$--$21$
points; on BPI-2017-Off it exceeds Neural Hawkes by $8.9$ points; on four of the BPI-2020 datasets and Train-Delays the gains are
$1$--$5$ points. The two exceptions are BPI-2015-2, where the
feature-stripped variant UMT\,$(\dagger)$ slightly outperforms UMT, and
BPI-2020-Dom, where IFTPP matches our model. Neural
Hawkes is the most competitive baseline overall. Transformer Hawkes
and GRU frequently collapse to near-zero accuracy on datasets with large
activity vocabularies (BPI-2015, Chicago-Food), suggesting
that these architectures do not scale well with the number of activities.

\paragraph{Duration prediction}
On MAE, UMT is best on $9$ of $13$ datasets. The largest
relative improvements appear on datasets whose inter-event times span multiple orders
of magnitude: Medical-Appts ($-15\%$ vs.\ Neural Hawkes),
Chicago-Food ($-27\%$), BPI-2013 ($-36\%$), and BPI-2017-Off
($-32\%$). Neural
Hawkes achieves the lowest MAE on the two BPI-2015 municipalities and
slightly outperforms UMT on BPI-2012 ($7.23$ vs.\ $7.56$ hours). GRU
achieves the lowest MAE on National-Grid ($0.02$), a
nearly periodic dataset with fixed hourly spacing where a constant
predictor already performs well. IFTPP is excluded from the MAE table because
it produced numerically divergent time predictions on $11$ of $13$ datasets.

\paragraph{Effect of multimodal features}
Comparing the full model against the feature-stripped variant
UMT\,$(\dagger)$ within the same table shows the contribution of
our multimodal feature encoder. On metadata-rich datasets the gap is
large: $+16$ accuracy points on BPI-2013, $+34$ on
Chicago-Food, and $+5$ on Train-Delays. MAE drops by $15$--$33\%$ on Chicago-Food, Medical-Appts, and
BPI-2013. On datasets where the activity dynamics are already near-Markov
(e.g., BPI-2020-Dom, BPI-2015-2), features provide
little or no benefit. This is consistent with the detailed ablation in
Section~\ref{sec:ablation}.

\subsection{What the model attends to}
\label{sec:attention}

The preceding results establish when context and sequence modeling improve
prediction. We next inspect the attention distributions used by the trained
model. The Perceiver produces one distribution over context fields for each
event, while each causal Transformer layer distributes attention over the
observed prefix. We record both distributions for the four ablation datasets
over three seeds. These models reproduce the corresponding accuracy results in
Table~\ref{tab:abl-features} within $0.3$ points.

\begin{figure}[t]
\centering
\begin{tikzpicture}[x=0.50cm, y=0.50cm, font=\small]
  \fill[heathi!6] (0,0) rectangle ++(1,-1); \fill[heathi!1] (1,0) rectangle ++(1,-1); \fill[heathi!14] (2,0) rectangle ++(1,-1); \fill[heathi!1] (3,0) rectangle ++(1,-1); \fill[heathi!16] (4,0) rectangle ++(1,-1); \fill[heathi!50] (5,0) rectangle ++(1,-1); \fill[heathi!1] (6,0) rectangle ++(1,-1); \fill[heathi!72] (7,0) rectangle ++(1,-1); \fill[heathi!39] (8,0) rectangle ++(1,-1); \fill[heathi!1] (9,0) rectangle ++(1,-1); \fill[heathi!2] (10,0) rectangle ++(1,-1); \fill[heathi!18] (11,0) rectangle ++(1,-1);
  \fill[heathi!11] (0,-1) rectangle ++(1,-1); \fill[heathi!9] (1,-1) rectangle ++(1,-1); \fill[heathi!65] (2,-1) rectangle ++(1,-1); \fill[heathi!5] (3,-1) rectangle ++(1,-1); \fill[heathi!51] (4,-1) rectangle ++(1,-1); \fill[heathi!48] (5,-1) rectangle ++(1,-1); \fill[heathi!9] (6,-1) rectangle ++(1,-1); \fill[heathi!13] (7,-1) rectangle ++(1,-1); \fill[heathi!17] (8,-1) rectangle ++(1,-1); \fill[heathi!3] (9,-1) rectangle ++(1,-1); \fill[heathi!6] (10,-1) rectangle ++(1,-1); \fill[heathi!8] (11,-1) rectangle ++(1,-1);
  \fill[heathi!15] (0,-2) rectangle ++(1,-1); \fill[heathi!8] (1,-2) rectangle ++(1,-1); \fill[heathi!13] (2,-2) rectangle ++(1,-1); \fill[heathi!1] (3,-2) rectangle ++(1,-1); \fill[heathi!23] (4,-2) rectangle ++(1,-1); \fill[heathi!51] (5,-2) rectangle ++(1,-1); \fill[heathi!2] (6,-2) rectangle ++(1,-1); \fill[heathi!63] (7,-2) rectangle ++(1,-1); \fill[heathi!48] (8,-2) rectangle ++(1,-1); \fill[heathi!2] (9,-2) rectangle ++(1,-1); \fill[heathi!2] (10,-2) rectangle ++(1,-1); \fill[heathi!12] (11,-2) rectangle ++(1,-1);
  \fill[heathi!4] (0,-3) rectangle ++(1,-1); \fill[heathi!4] (1,-3) rectangle ++(1,-1); \fill[heathi!3] (2,-3) rectangle ++(1,-1); \fill[heathi!4] (3,-3) rectangle ++(1,-1); \fill[heathi!15] (4,-3) rectangle ++(1,-1); \fill[heathi!7] (5,-3) rectangle ++(1,-1); \fill[heathi!4] (6,-3) rectangle ++(1,-1); \fill[heathi!97] (7,-3) rectangle ++(1,-1); \fill[heathi!10] (8,-3) rectangle ++(1,-1); \fill[heathi!4] (9,-3) rectangle ++(1,-1); \fill[heathi!3] (10,-3) rectangle ++(1,-1); \fill[heathi!2] (11,-3) rectangle ++(1,-1);
  \foreach \x in {0,...,11} {\foreach \y in {0,...,3} {\draw[white, very thin] (\x,-\y) rectangle ++(1,-1);}}
  \node[anchor=east] at (-0.2,-0.5) {\textsf{1 Create Offer}};
  \node[anchor=east] at (-0.2,-1.5) {\textsf{2 Created}};
  \node[anchor=east] at (-0.2,-2.5) {\textsf{3 Sent}};
  \node[anchor=east] at (-0.2,-3.5) {\textsf{4 Returned}};
  \node[rotate=90, anchor=west] at (0.5,0.2) {\textsf{Action}};
  \node[rotate=90, anchor=west] at (1.5,0.2) {\textsf{Activity}};
  \node[rotate=90, anchor=west] at (2.5,0.2) {\textsf{Resource}};
  \node[rotate=90, anchor=west] at (3.5,0.2) {\textsf{Origin}};
  \node[rotate=90, anchor=west] at (4.5,0.2) {\textsf{Accepted}};
  \node[rotate=90, anchor=west] at (5.5,0.2) {\textsf{Selected}};
  \node[rotate=90, anchor=west] at (6.5,0.2) {\textsf{Lifecycle}};
  \node[rotate=90, anchor=west] at (7.5,0.2) {\textsf{CreditScore}};
  \node[rotate=90, anchor=west] at (8.5,0.2) {\textsf{1stWithdrawal}};
  \node[rotate=90, anchor=west] at (9.5,0.2) {\textsf{MonthlyCost}};
  \node[rotate=90, anchor=west] at (10.5,0.2) {\textsf{NumTerms}};
  \node[rotate=90, anchor=west] at (11.5,0.2) {\textsf{OfferedAmt}};
  \node[anchor=north] at (6,-4.5) {context field};
\end{tikzpicture}
\caption{Perceiver cross-attention for a four-event BPI-2017-Off prefix
(trace~\#3864).  Darker blue indicates higher attention weight.
The bottom row (event~4) is the position from which the model forecasts
the fifth event; there, the fusion concentrates $95\%$ of its mass on
\textsf{CreditScore}.}
\Description{A $4 \times 12$ blue-scale heat map mapping four prefix events
to twelve context fields.  The bottom row concentrates almost all mass on
the CreditScore column.}
\label{fig:attn-fusion}
\end{figure}
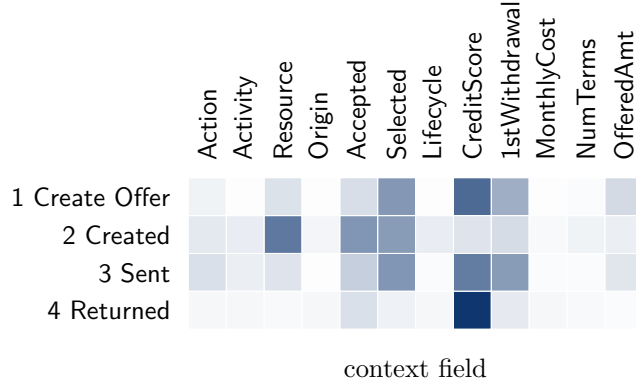

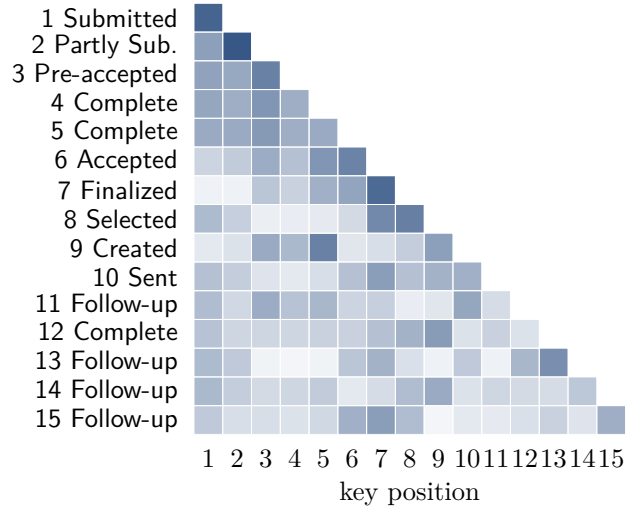
\begin{figure}[t]
\centering
\begin{tikzpicture}[x=0.38cm, y=0.38cm, font=\small]
  \fill[heathi!75] (0,0) rectangle ++(1,-1);
  \fill[heathi!46] (0,-1) rectangle ++(1,-1); \fill[heathi!81] (1,-1) rectangle ++(1,-1);
  \fill[heathi!45] (0,-2) rectangle ++(1,-1); \fill[heathi!42] (1,-2) rectangle ++(1,-1); \fill[heathi!61] (2,-2) rectangle ++(1,-1);
  \fill[heathi!43] (0,-3) rectangle ++(1,-1); \fill[heathi!39] (1,-3) rectangle ++(1,-1); \fill[heathi!51] (2,-3) rectangle ++(1,-1); \fill[heathi!39] (3,-3) rectangle ++(1,-1);
  \fill[heathi!41] (0,-4) rectangle ++(1,-1); \fill[heathi!41] (1,-4) rectangle ++(1,-1); \fill[heathi!49] (2,-4) rectangle ++(1,-1); \fill[heathi!39] (3,-4) rectangle ++(1,-1); \fill[heathi!41] (4,-4) rectangle ++(1,-1);
  \fill[heathi!21] (0,-5) rectangle ++(1,-1); \fill[heathi!25] (1,-5) rectangle ++(1,-1); \fill[heathi!40] (2,-5) rectangle ++(1,-1); \fill[heathi!30] (3,-5) rectangle ++(1,-1); \fill[heathi!51] (4,-5) rectangle ++(1,-1); \fill[heathi!60] (5,-5) rectangle ++(1,-1);
  \fill[heathi!7] (0,-6) rectangle ++(1,-1); \fill[heathi!7] (1,-6) rectangle ++(1,-1); \fill[heathi!28] (2,-6) rectangle ++(1,-1); \fill[heathi!22] (3,-6) rectangle ++(1,-1); \fill[heathi!38] (4,-6) rectangle ++(1,-1); \fill[heathi!44] (5,-6) rectangle ++(1,-1); \fill[heathi!72] (6,-6) rectangle ++(1,-1);
  \fill[heathi!33] (0,-7) rectangle ++(1,-1); \fill[heathi!23] (1,-7) rectangle ++(1,-1); \fill[heathi!8] (2,-7) rectangle ++(1,-1); \fill[heathi!9] (3,-7) rectangle ++(1,-1); \fill[heathi!10] (4,-7) rectangle ++(1,-1); \fill[heathi!18] (5,-7) rectangle ++(1,-1); \fill[heathi!57] (6,-7) rectangle ++(1,-1); \fill[heathi!62] (7,-7) rectangle ++(1,-1);
  \fill[heathi!11] (0,-8) rectangle ++(1,-1); \fill[heathi!14] (1,-8) rectangle ++(1,-1); \fill[heathi!41] (2,-8) rectangle ++(1,-1); \fill[heathi!34] (3,-8) rectangle ++(1,-1); \fill[heathi!61] (4,-8) rectangle ++(1,-1); \fill[heathi!12] (5,-8) rectangle ++(1,-1); \fill[heathi!16] (6,-8) rectangle ++(1,-1); \fill[heathi!24] (7,-8) rectangle ++(1,-1); \fill[heathi!46] (8,-8) rectangle ++(1,-1);
  \fill[heathi!30] (0,-9) rectangle ++(1,-1); \fill[heathi!24] (1,-9) rectangle ++(1,-1); \fill[heathi!13] (2,-9) rectangle ++(1,-1); \fill[heathi!11] (3,-9) rectangle ++(1,-1); \fill[heathi!16] (4,-9) rectangle ++(1,-1); \fill[heathi!30] (5,-9) rectangle ++(1,-1); \fill[heathi!47] (6,-9) rectangle ++(1,-1); \fill[heathi!30] (7,-9) rectangle ++(1,-1); \fill[heathi!37] (8,-9) rectangle ++(1,-1); \fill[heathi!38] (9,-9) rectangle ++(1,-1);
  \fill[heathi!32] (0,-10) rectangle ++(1,-1); \fill[heathi!19] (1,-10) rectangle ++(1,-1); \fill[heathi!40] (2,-10) rectangle ++(1,-1); \fill[heathi!29] (3,-10) rectangle ++(1,-1); \fill[heathi!35] (4,-10) rectangle ++(1,-1); \fill[heathi!21] (5,-10) rectangle ++(1,-1); \fill[heathi!23] (6,-10) rectangle ++(1,-1); \fill[heathi!9] (7,-10) rectangle ++(1,-1); \fill[heathi!12] (8,-10) rectangle ++(1,-1); \fill[heathi!43] (9,-10) rectangle ++(1,-1); \fill[heathi!17] (10,-10) rectangle ++(1,-1);
  \fill[heathi!29] (0,-11) rectangle ++(1,-1); \fill[heathi!20] (1,-11) rectangle ++(1,-1); \fill[heathi!20] (2,-11) rectangle ++(1,-1); \fill[heathi!20] (3,-11) rectangle ++(1,-1); \fill[heathi!22] (4,-11) rectangle ++(1,-1); \fill[heathi!22] (5,-11) rectangle ++(1,-1); \fill[heathi!30] (6,-11) rectangle ++(1,-1); \fill[heathi!37] (7,-11) rectangle ++(1,-1); \fill[heathi!48] (8,-11) rectangle ++(1,-1); \fill[heathi!14] (9,-11) rectangle ++(1,-1); \fill[heathi!22] (10,-11) rectangle ++(1,-1); \fill[heathi!14] (11,-11) rectangle ++(1,-1);
  \fill[heathi!33] (0,-12) rectangle ++(1,-1); \fill[heathi!26] (1,-12) rectangle ++(1,-1); \fill[heathi!6] (2,-12) rectangle ++(1,-1); \fill[heathi!5] (3,-12) rectangle ++(1,-1); \fill[heathi!6] (4,-12) rectangle ++(1,-1); \fill[heathi!28] (5,-12) rectangle ++(1,-1); \fill[heathi!37] (6,-12) rectangle ++(1,-1); \fill[heathi!15] (7,-12) rectangle ++(1,-1); \fill[heathi!7] (8,-12) rectangle ++(1,-1); \fill[heathi!26] (9,-12) rectangle ++(1,-1); \fill[heathi!7] (10,-12) rectangle ++(1,-1); \fill[heathi!35] (11,-12) rectangle ++(1,-1); \fill[heathi!54] (12,-12) rectangle ++(1,-1);
  \fill[heathi!34] (0,-13) rectangle ++(1,-1); \fill[heathi!24] (1,-13) rectangle ++(1,-1); \fill[heathi!18] (2,-13) rectangle ++(1,-1); \fill[heathi!20] (3,-13) rectangle ++(1,-1); \fill[heathi!25] (4,-13) rectangle ++(1,-1); \fill[heathi!11] (5,-13) rectangle ++(1,-1); \fill[heathi!17] (6,-13) rectangle ++(1,-1); \fill[heathi!32] (7,-13) rectangle ++(1,-1); \fill[heathi!41] (8,-13) rectangle ++(1,-1); \fill[heathi!14] (9,-13) rectangle ++(1,-1); \fill[heathi!20] (10,-13) rectangle ++(1,-1); \fill[heathi!18] (11,-13) rectangle ++(1,-1); \fill[heathi!17] (12,-13) rectangle ++(1,-1); \fill[heathi!27] (13,-13) rectangle ++(1,-1);
  \fill[heathi!26] (0,-14) rectangle ++(1,-1); \fill[heathi!16] (1,-14) rectangle ++(1,-1); \fill[heathi!16] (2,-14) rectangle ++(1,-1); \fill[heathi!14] (3,-14) rectangle ++(1,-1); \fill[heathi!19] (4,-14) rectangle ++(1,-1); \fill[heathi!38] (5,-14) rectangle ++(1,-1); \fill[heathi!47] (6,-14) rectangle ++(1,-1); \fill[heathi!32] (7,-14) rectangle ++(1,-1); \fill[heathi!5] (8,-14) rectangle ++(1,-1); \fill[heathi!11] (9,-14) rectangle ++(1,-1); \fill[heathi!10] (10,-14) rectangle ++(1,-1); \fill[heathi!15] (11,-14) rectangle ++(1,-1); \fill[heathi!22] (12,-14) rectangle ++(1,-1); \fill[heathi!13] (13,-14) rectangle ++(1,-1); \fill[heathi!39] (14,-14) rectangle ++(1,-1);
  \foreach \row in {0,...,14} {\foreach \col in {0,...,\row} {\draw[white, very thin] (\col,-\row) rectangle ++(1,-1);}}
  \node[anchor=east] at (-0.2,-0.5) {\textsf{1 Submitted}};
  \node[anchor=east] at (-0.2,-1.5) {\textsf{2 Partly Sub.}};
  \node[anchor=east] at (-0.2,-2.5) {\textsf{3 Pre-accepted}};
  \node[anchor=east] at (-0.2,-3.5) {\textsf{4 Complete}};
  \node[anchor=east] at (-0.2,-4.5) {\textsf{5 Complete}};
  \node[anchor=east] at (-0.2,-5.5) {\textsf{6 Accepted}};
  \node[anchor=east] at (-0.2,-6.5) {\textsf{7 Finalized}};
  \node[anchor=east] at (-0.2,-7.5) {\textsf{8 Selected}};
  \node[anchor=east] at (-0.2,-8.5) {\textsf{9 Created}};
  \node[anchor=east] at (-0.2,-9.5) {\textsf{10 Sent}};
  \node[anchor=east] at (-0.2,-10.5) {\textsf{11 Follow-up}};
  \node[anchor=east] at (-0.2,-11.5) {\textsf{12 Complete}};
  \node[anchor=east] at (-0.2,-12.5) {\textsf{13 Follow-up}};
  \node[anchor=east] at (-0.2,-13.5) {\textsf{14 Follow-up}};
  \node[anchor=east] at (-0.2,-14.5) {\textsf{15 Follow-up}};
  \foreach \x in {1,...,15} {\node[anchor=north] at (\x-0.5,-15.2) {\x};}
  \node[anchor=north] at (7.5,-16.3) {key position};
\end{tikzpicture}
\caption{Layer-4 causal self-attention for a 15-event BPI-2012 prefix
(trace~\#1097); color scale as in Figure~\ref{fig:attn-fusion}.
Attention patterns vary by position; the bottom row (event~15) spreads
mass across the prefix, reaching back to semantically relevant
application-finalization events rather than attending only locally.}
\Description{A $15 \times 15$ lower-triangular blue-scale heat map for a
BPI-2012 prefix.  The bottom row spreads mass broadly with peaks at
distant events.}
\label{fig:attn-backbone}
\end{figure}

\paragraph{Perceiver cross-attention.}
Figure~\ref{fig:attn-fusion} shows the Perceiver cross-attention matrix for a
four-event BPI-2017-Off prefix: rows are the four observed events, columns are
the twelve context fields available at each event.  These include the activity
name, the handling employee (\textsf{Resource}), the event subsystem
(\textsf{Origin}: Application, Offer, or Workflow), the task state
(\textsf{Lifecycle}: e.g.\ complete, start), and several case-level loan
attributes.  The bottom row is the prediction position from which the model
forecasts the fifth event.  There, the Perceiver assigns $95\%$ of its mass to
\textsf{CreditScore} and predicts \emph{Accepted} with probability
$1.00$.  Replacing only this value with one from another case reduces that
probability to zero and changes the prediction to \emph{Refused}.

This pattern generalizes beyond a single trace.  Across the four ablation
datasets, the rank correlation between each column's mean attention mass
and its mutual information (MI) with the next activity is positive
(Spearman $\rho = {+}0.37$ to ${+}0.63$), and on BPI-2012 all five of the five
highest-MI columns are among the five most attended.  The Perceiver thus
recovers an MI-like ranking without ever being trained on it.  Yet it also
departs from MI in one systematic way: the activity name, which has the
highest MI of any column, receives \emph{less} than chance attention
because the backbone already receives the activity as its input token.
A static MI-based selection would retain this redundant feature; the
Perceiver learns to ignore it and spend its capacity on columns whose
information is not already in the token stream.  On datasets where context
features are uninformative (BPI-2015), the fusion stays near-uniform
rather than forcing a selection, matching the ablation findings of
Table~\ref{tab:abl-features}.

We validate these attention patterns with a causal intervention: for each
of $48$ high-confidence prediction positions per dataset, we replace one
attribute at a time (i.e., one column in Figure~\ref{fig:attn-fusion}, such
as \textsf{CreditScore} or \textsf{Resource}) with the corresponding value
from a different case and measure the drop in predicted probability.  If
attention were uninformative, the most-attended attribute would coincide with
the most damaging one only at the chance rate of $1/N$ ($20\%$ for $5$ fields
on BPI-2012, $8.3\%$ for $12$ on BPI-2017-Off).  In practice the agreement
is $94\%$ and $83\%$, respectively, and replacing the most-attended field
drops the predicted probability by $0.79$ and $0.34$ on average, while
replacing a median-attended field has no measurable effect.  On the two
BPI-2015 logs, where context contributes little, the correspondence is
weaker but still above chance.

\paragraph{Backbone attention.}
Figure~\ref{fig:attn-backbone} shows the full lower-triangular
self-attention matrix from the final layer (layer~4 of~4) of the causal
Transformer backbone for a 15-event BPI-2012 prefix.
Row~$i$ records how query~$i$ distributes attention over keys $\leq i$.
The matrix reveals position-dependent patterns: some events attend
primarily to their immediate predecessors, while others spread mass across
the full observed prefix.  Notably, the prediction row (row~15) reaches
back to semantically relevant events such as the two
application-finalization steps rather than relying on local context alone.
Aggregated across all prediction positions, the three long-trace datasets
place $47$--$59\%$ of backbone mass on events at least four positions back,
confirming that the model is not a Markov reader.

\subsection{Ablation Study}
\label{sec:ablation}

The main results establish that UMT outperforms existing baselines;
we now investigate the contribution of each component. Specifically: (1)~Does the multi-scale inter-arrival
head improve time prediction over standard point-process likelihood heads,
and at what cost to activity accuracy? (2)~Does the universal feature
encoder improve performance, and which modalities carry the signal?
(3)~Does the Perceiver fusion mechanism enable scaling to many attributes
without degradation? (4)~Are the results sensitive to model capacity or
hyperparameter choice?

Each ablation changes \emph{one} design decision against the fixed
reference configuration described in Section~\ref{sec:baselines}. All ablations are trained and evaluated under the temporal-split protocol of
Section~\ref{sec:setup}. We run every ablation on four datasets
(Table~\ref{tab:datasets}), chosen to span two regimes.
BPI-2017-Off is metadata-dominated ($12$ attribute columns, $7$ categorical and
$5$ numerical, with only $8$ activity types; $31{,}063$/$6{,}450$
train/test cases). The two BPI-2015 municipalities are the opposite extreme
(${\approx}400$ activity types and $23$--$24$ columns over fewer than $900$
training traces). BPI-2012 lies in between ($24$ activities, $5$ columns,
$9{,}454$/$1{,}964$ train/test). Averaging over all datasets would hide
this contrast.

In all
tables below, positive $\Delta$accuracy means the \emph{ablation} is better, and
positive $\Delta$MAE means the ablation is \emph{worse}.

\begin{table}[t]
\caption{Feature-modality ablation: next-activity accuracy (mean over five
seeds). The first row is our full model in absolute percent; every row below
is the change when the input is restricted to one modality. Positive
$\Delta$accuracy means the restriction is \emph{better}.}
\medskip
\label{tab:abl-features}
\centering
\footnotesize
\setlength{\tabcolsep}{4pt}
\begin{tabular}{@{}lrrrr@{}}
\toprule
Restriction & BPI-2012 & BPI-2015-1 & BPI-2015-2 & BPI-2017-Off \\
\midrule
\textbf{Ours (\%)} & $\mathbf{86.03}$ & $\mathbf{51.03}$ & $\mathbf{50.29}$ & $\mathbf{92.41}$ \\
\midrule
No context       & $-1.90$ & $+1.02$ & $+3.82$ & $-17.95$ \\
Categorical only & $-0.07$ & $+0.16$ & $-0.73$ & $-2.94$ \\
Numerical only   & $-1.97$ & $+0.95$ & $+1.73$ & $-7.17$ \\
Textual only     & $-2.07$ & $+0.28$ & $+1.32$ & $-18.10$ \\
Case-level only  & $-1.60$ & $+0.33$ & $-0.04$ & $-0.00$ \\
Event-level only & $-0.46$ & $+0.50$ & $+0.26$ & $-18.13$ \\
\bottomrule
\end{tabular}
\end{table}

\begin{table}[t]
\caption{Feature-modality ablation: inter-arrival MAE (same runs and
reading convention as Table~\ref{tab:abl-features}; positive $\Delta$MAE
means the ablation is \emph{worse}). Units: hours for BPI-2012/2015-1/2015-2,
days for BPI-2017-Off.}
\medskip
\label{tab:abl-features-mae}
\centering
\footnotesize
\setlength{\tabcolsep}{4pt}
\begin{tabular}{@{}lrrrr@{}}
\toprule
Restriction & BPI-2012 & BPI-2015-1 & BPI-2015-2 & BPI-2017-Off \\
\midrule
\textbf{Ours} & $\mathbf{7.93}$ & $\mathbf{28.74}$ & $\mathbf{31.36}$ & $\mathbf{3.71}$ \\
\midrule
No context       & $+0.19$ & $-0.50$ & $-1.99$ & $+0.79$ \\
Categorical only & $-0.04$ & $+0.05$ & $+3.31$ & $+0.01$ \\
Numerical only   & $+0.19$ & $-1.92$ & $-2.22$ & $+0.44$ \\
Textual only     & $+0.28$ & $-1.23$ & $-0.79$ & $+0.79$ \\
Case-level only  & $+0.40$ & $-1.33$ & $+1.58$ & $-0.02$ \\
Event-level only & $-0.16$ & $-0.29$ & $+0.12$ & $+0.83$ \\
\bottomrule
\end{tabular}
\end{table}

\paragraph{Does the universal feature encoder improve performance, and which modalities
carry the signal?}
Is encoding heterogeneous case and event attributes beneficial? And is the
four-way treatment (categorical\,/\,numerical\,/\,textual $\times$
case-level\,/\,event-level) of Section~\ref{sec:feature-encoder} necessary, or
is a single modality sufficient? Table~\ref{tab:abl-features} restricts the input to one
modality at a time. No single modality is sufficient, and which one matters
varies across datasets. On BPI-2017-Off the signal is entirely case-level and
non-textual. Case attributes alone reproduce the full model ($-0.00$), while event
attributes alone drop to the no-feature baseline ($-18.13$). Categorical
or numerical alone recover only $15$ and $11$ of the $18$ points. On
BPI-2012 the ordering reverses and event-level attributes are the best single
restriction ($-0.46$) while numerical or textual alone lose ${\approx}2$ points.
Every restriction costs accuracy somewhere. The MAE panel shows that these
activity effects do not trade off against time prediction: removing all context
features raises BPI-2017-Off MAE by $0.79$ days ($21\%$), whereas it reduces
MAE by $1.7\%$ and $6.3\%$ on BPI-2015-1/2, the two logs where removing context
also improves accuracy. The shared encoder also feeds the time head
(Table~\ref{tab:mae}: $81.5\to121.3$ days on Chicago-Food without
features). On the two BPI-2015 municipalities, however,
features are neutral-to-harmful ($+1.0$ and $+3.8$ points without them),
because $23$--$24$ columns over fewer than $900$ traces with ${\approx}400$
activity types present primarily an overfitting risk;
Table~\ref{tab:accuracy} shows the same pattern across all $16$ datasets. The useful modality is
dataset-dependent and no single one suffices. A universal encoder that ingests all
four kinds without per-dataset engineering is therefore well-motivated.

\begin{table*}[t]
\caption{Scaling to a fixed context width. Each log is padded with synthetic,
label-independent \emph{distractor} columns until it has exactly $100$ context
columns; the left number of each pair is the log's own attribute count, the right
one is $100$. The two panels differ only in the modality of the padding:
high-cardinality categorical columns and standardised numerical ones.
Next-activity accuracy in percent, mean over three seeds, all other
settings as in the reference configuration. The last row is the same model with
every context column switched off.}
\label{tab:abl-scaling}
\medskip
\centering
\footnotesize
\setlength{\tabcolsep}{5pt}
\begin{tabular}{@{}lrrrrrrrr@{}}
\toprule
 & \multicolumn{2}{c}{BPI-2012} & \multicolumn{2}{c}{BPI-2015-1} & \multicolumn{2}{c}{BPI-2015-2} & \multicolumn{2}{c}{BPI-2017-Off} \\
\cmidrule(lr){2-3}\cmidrule(lr){4-5}\cmidrule(lr){6-7}\cmidrule(lr){8-9}
\#\,context columns & $5$ & $100$ & $23$ & $100$ & $24$ & $100$ & $12$ & $100$ \\
\midrule
\multicolumn{9}{@{}l}{\emph{Categorical distractor columns}} \\
  Additive sum & $85.89$ & $85.12$ & $52.06$ & $40.45$ & $50.13$ & $36.62$ & $92.38$ & $92.21$ \\
  Perceiver, time query & $85.70$ & $85.94$ & $51.21$ & $50.88$ & $46.28$ & $46.23$ & $92.41$ & $92.39$ \\
  Perceiver, act.+time query \textbf{(ours)} & $86.09$ & $85.42$ & $51.18$ & $49.61$ & $51.65$ & $49.24$ & $92.46$ & $92.37$ \\
\midrule
\multicolumn{9}{@{}l}{\emph{Numerical distractor columns}} \\
  Additive sum & $85.89$ & $85.80$ & $52.06$ & $50.90$ & $50.13$ & $47.52$ & $92.38$ & $92.25$ \\
  Perceiver, time query & $85.70$ & $85.84$ & $51.21$ & $51.04$ & $46.28$ & $45.56$ & $92.41$ & $92.33$ \\
  Perceiver, act.+time query \textbf{(ours)} & $86.09$ & $85.53$ & $51.18$ & $48.87$ & $51.65$ & $48.08$ & $92.46$ & $92.37$ \\
\midrule
  No context features at all & $84.11$ & $84.11$ & $52.12$ & $52.12$ & $53.56$ & $53.56$ & $74.36$ & $74.36$ \\
\bottomrule
\end{tabular}
\end{table*}

\paragraph{Does the Perceiver fusion support scaling to many attributes?}
The encoder accepts any number of heterogeneous columns; does accuracy survive as
that number grows? No log contains hundreds of \emph{informative} attributes, so we
pad each of the four ablation logs with synthetic, label-independent
\emph{distractor} columns until every log carries exactly $100$ context columns
(Table~\ref{tab:abl-scaling}) --- a $4\times$ widening for the two municipal logs
and a $20\times$ one for BPI-2012. The experiment is run twice, once with
high-cardinality categorical padding and once with standardised numerical padding,
because the two modalities enter the encoder differently. Every column is drawn
independently of the label, so the ideal behaviour is to be unchanged. As a
comparison, \emph{additive fusion} directly sums all encoded field vectors and
combines that aggregate with the activity and time embeddings; unlike the
Perceiver, it applies no event-conditioned weighting or normalization across
fields.

Over the eight padding conditions per fusion variant (four logs $\times$ two
modalities), additive summation loses up to $13.5$ accuracy points, whereas the
time-query Perceiver loses at most $0.72$ points, with a median loss of $0.07$
points. The activity-and-time-query configuration used in the main tables loses
at most $3.6$ points. Additive fusion is not merely diluted by padding:
with categorical padding, it falls to $40.5$ and $36.6$ on the two municipal logs,
$11.7$ and $16.9$ points below its own no-feature baseline. Both Perceiver
variants remain within $0.8$ points of their native-width accuracy on BPI-2012
and BPI-2017-Off, the two logs whose context features carry signal.

\begin{table}[t]
\caption{Inter-arrival time head. First row: our head, absolute MAE in days;
every row below is the change in MAE (percent) when that head is swapped in on
the same backbone, features and training budget.
Next-activity accuracy is within $1.1$ points of the reference throughout.}
\medskip
\label{tab:abl-timehead}
\centering
\footnotesize
\setlength{\tabcolsep}{3pt}
\begin{tabular}{@{}lrrrr@{}}
\toprule
Time head & BPI-2012 & BPI-2015-1 & BPI-2015-2 & BPI-2017-Off \\
\midrule
\textbf{Ours: multi-scale (d)} & $\mathbf{0.331}$ & $\mathbf{1.208}$ & $\mathbf{1.326}$ & $\mathbf{3.728}$ \\
\midrule
LogNormal                       & $+6.8$ & $+14.1$ & $+23.4$          & $+4.3$ \\
Mixture of LogNormal $k{=}1$          & $+6.5$ & $+14.1$ & $+27.6$          & $+7.2$ \\
Mixture of LogNormal $k{=}3$          & $+7.1$ & $+9.7$  & $+17.6$          & $+5.1$ \\
Mixture of LogNormal $k{=}5$          & $+6.6$ & $+9.2$  & $+21.4$          & $+4.8$ \\
Poisson day count               & $+5.8$ & $+20.8$ & $\mathbf{+52.7}$ & $+4.3$ \\
\bottomrule
\end{tabular}
\end{table}

\paragraph{Multi-scale inter-arrival head versus standard point-process heads.}
Is the day-count plus within-day decomposition of
Section~\ref{sec:multiscale-time} better than the log-normal and
mixture-log-normal likelihood heads used by intensity-free temporal point-process
models? And does it cost activity accuracy?
Table~\ref{tab:abl-timehead} swaps the head and leaves everything else fixed.
Next-activity accuracy remains within $1.1$ points of the reference for every head variant (only the
five-component mixture drops by $1.05$ points). The
comparison is therefore about time. There the decomposition wins on \emph{every}
dataset against \emph{every} baseline head, by $9.9\%$ to $13.8\%$ MAE.
The largest gaps appear on the two long-tailed
BPI-2015 datasets (up to $+28\%$). Additional mixture components do not close this gap. The
deficit is structural rather than due to limited capacity. Over-dispersion is the
key factor: replacing the negative-binomial day count
of Equation~\eqref{eq:day-distribution} by a Poisson costs $20.9\%$ MAE
and $52.7\%$ on BPI-2015-2
($1.33\to2.00$ days). The multi-scale head thus produces the duration results
of Table~\ref{tab:mae} at no cost in accuracy. Its over-dispersed day count
is essential.

\paragraph{Model capacity and hyperparameter sensitivity.}
Are the reported results an artifact of model capacity or hyperparameter
choice? We vary width ($d_{\text{model}}\in\{16,32,64,128\}$), depth
($2$--$8$ layers), learning rate ($10^{-4}$ to $3{\cdot}10^{-3}$), and weight
decay ($0$ to $0.1$) around the reference configuration. Performance is
insensitive to these choices except at the extremes of the grid. The
sensitivity that remains is confined to the two BPI-2015 datasets with
${\approx}400$ activity types. Reducing capacity degrades performance
($d_{\text{model}}=16$: $-9.8$ and $-11.2$ points on BPI-2015-1/2;
$d_{\text{model}}=32$: $-4.6$; $2$ layers:
$-2.3$), as does reducing the learning rate ($10^{-4}$:
${\approx}-2$). Every other setting stays within $1.5$ points of the reference
on every dataset. MAE spans just $-1.8\%$ to
$+4.9\%$ across the whole grid. Increasing the width does not help
($d_{\text{model}}=128$: $+0.49$ points at four times the
parameters, with $4.9\%$ worse MAE). The reference configuration therefore lies on a performance plateau
rather than at a tuned optimum. No setting in this grid yields more than half
an accuracy point or $2\%$ MAE improvement, well below the architectural
effects reported above ($1.6$--$18$ points, $10$--$55\%$ MAE).

\section{Conclusion}

We presented Universal Multi-Modal Traceformer (UMT), a unified framework for next-event prediction that integrates heterogeneous contextual information from event logs. UMT combines three contributions: a universal feature encoder that maps categorical, numerical, and textual fields into a shared latent space; a per-event Perceiver module that adaptively weights contextual features through event-conditioned cross-attention; and a multi-scale inter-arrival time representation that decomposes intervals into an overdispersed day count and cyclic within-day components.

Experiments on 13 real-world event logs (spanning business workflows, IT incident management, healthcare, food safety, and rail operations) show that UMT achieves the highest next-activity accuracy on 11 of 13 datasets and the lowest duration MAE on 9 of 13. Ablations confirm that each component contributes distinctly: the multi-scale time head reduces MAE by 12\% on average over a log-normal head, and multimodal features add up to 18 accuracy points on attribute-rich logs. The Perceiver-based fusion degrades 6.8$\times$ more slowly than additive alternatives as the number of input attributes grows, removing the need for per-dataset feature engineering. Together, these results demonstrate that a single architecture can ingest arbitrary combinations of event- and trace-level metadata across domains without hand-crafted feature pipelines, addressing a key obstacle to deploying process prediction in heterogeneous enterprise environments.

Several directions remain. First, UMT currently predicts one step ahead; extending it to multi-step or remaining-time forecasting would cover additional process-monitoring tasks common in operations research and service management. Second, the universal encoder handles categorical, numerical, and textual fields; incorporating additional modalities such as images, audio, or graph-structured inter-case relations would broaden its applicability. Third, the Perceiver cross-attention weights provide per-feature importance scores at each event position that could support interpretability requirements in regulated domains such as healthcare and finance.

\section*{Ethical Considerations}

All datasets used in this work are publicly available and anonymized by their providers. Event logs may contain sensitive attributes; deployments on proprietary logs should include appropriate access controls and privacy safeguards. Because the model ingests all available features, practitioners should audit inputs for personally identifiable information. Next-event predictions in domains such as loan processing or healthcare could reproduce biases present in the training data and should support, not replace, human decisions. All experiments ran on a single NVIDIA L40S GPU with a total compute budget of under 100 GPU-hours. Code and preprocessing scripts will be released upon acceptance to facilitate reproducibility. We encourage practitioners to perform bias audits and fairness evaluations before deploying predictions from UMT or similar models in high-stakes decision pipelines. No personal data was collected or generated by this study.

% \clearpage
\bibliographystyle{ACM-Reference-Format}
\bibliography{references}


\begin{thebibliography}{45}

%%% the Web address.

\ifx \showCODEN    \undefined \def \showCODEN     #1{\unskip}     \fi
\ifx \showISBNx    \undefined \def \showISBNx     #1{\unskip}     \fi
\ifx \showISBNxiii \undefined \def \showISBNxiii  #1{\unskip}     \fi
\ifx \showISSN     \undefined \def \showISSN      #1{\unskip}     \fi
\ifx \showLCCN     \undefined \def \showLCCN      #1{\unskip}     \fi
\ifx \shownote     \undefined \def \shownote      #1{#1}          \fi
\ifx \showarticletitle \undefined \def \showarticletitle #1{#1}   \fi
\ifx \showURL      \undefined \def \showURL       {\relax}        \fi
% invisible to TeX
\providecommand\bibfield[2]{#2}
\providecommand\bibinfo[2]{#2}
\providecommand\natexlab[1]{#1}
\providecommand\showeprint[2][]{arXiv:#2}

\bibitem[dat(2025)]%
        {dataset_itincidents}
 \bibinfo{year}{2025}\natexlab{}.
\newblock \bibinfo{title}{{IT} Incidents Event Log}.
\newblock \bibinfo{howpublished}{Zenodo,
  \url{https://zenodo.org/records/15484999}}.
\newblock


\bibitem[{4TU.ResearchData}(2020)]%
        {bpichallenge}
\bibfield{author}{\bibinfo{person}{{4TU.ResearchData}}.}
  \bibinfo{year}{2012--2020}\natexlab{}.
\newblock \bibinfo{title}{{Business Process Intelligence Challenge} event
  logs}.
\newblock \bibinfo{howpublished}{4TU.ResearchData, \url{https://data.4tu.nl/}}.
\newblock
\newblock
\shownote{Event-log collection; each yearly challenge is registered under its
  own DOI (prefix 10.4121)}.


\bibitem[Badami(2018)]%
        {dataset_traindelays}
\bibfield{author}{\bibinfo{person}{Pranav Badami}.}
  \bibinfo{year}{2018}\natexlab{}.
\newblock \bibinfo{title}{{NJ} Transit {Amtrak} {NEC} Performance}.
\newblock \bibinfo{howpublished}{Kaggle,
  \url{https://www.kaggle.com/datasets/pranavbadami/nj-transit-amtrak-nec-performance}}.
\newblock


\bibitem[Blundell et~al\mbox{.}(2012)]%
        {blundell2012modelling}
\bibfield{author}{\bibinfo{person}{Charles Blundell}, \bibinfo{person}{Jeff
  Beck}, {and} \bibinfo{person}{Katherine~A Heller}.}
  \bibinfo{year}{2012}\natexlab{}.
\newblock \showarticletitle{Modelling reciprocating relationships with Hawkes
  processes}. In \bibinfo{booktitle}{\emph{Advances in Neural Information
  Processing Systems}}. \bibinfo{pages}{2600--2608}.
\newblock


\bibitem[Bukhsh et~al\mbox{.}(2021)]%
        {bukhsh2021processtransformer}
\bibfield{author}{\bibinfo{person}{Zaharah~Allah Bukhsh},
  \bibinfo{person}{Aaqib Saeed}, {and} \bibinfo{person}{Remco~M. Dijkman}.}
  \bibinfo{year}{2021}\natexlab{}.
\newblock \showarticletitle{ProcessTransformer: Predictive Business Process
  Monitoring with Transformer Network}.
\newblock \bibinfo{journal}{\emph{CoRR}}  \bibinfo{volume}{abs/2104.00721}
  (\bibinfo{year}{2021}).
\newblock


\bibitem[Camargo et~al\mbox{.}(2019)]%
        {camargo2019learning}
\bibfield{author}{\bibinfo{person}{Manuel Camargo}, \bibinfo{person}{Marlon
  Dumas}, {and} \bibinfo{person}{Oscar Gonz{\'a}lez-Rojas}.}
  \bibinfo{year}{2019}\natexlab{}.
\newblock \showarticletitle{Learning Accurate {LSTM} Models of Business
  Processes}. In \bibinfo{booktitle}{\emph{Business Process Management (BPM)}}
  \emph{(\bibinfo{series}{Lecture Notes in Computer Science},
  Vol.~\bibinfo{volume}{11675})}. \bibinfo{publisher}{Springer},
  \bibinfo{pages}{286--302}.
\newblock
\href{https://doi.org/10.1007/978-3-030-26619-6_19}{doi:\nolinkurl{10.1007/978-3-030-26619-6_19}}


\bibitem[Charlin et~al\mbox{.}(2015)]%
        {charlin2015dynamic}
\bibfield{author}{\bibinfo{person}{Laurent Charlin}, \bibinfo{person}{Rajesh
  Ranganath}, \bibinfo{person}{James McInerney}, {and} \bibinfo{person}{David~M
  Blei}.} \bibinfo{year}{2015}\natexlab{}.
\newblock \showarticletitle{Dynamic {P}oisson factorization}. In
  \bibinfo{booktitle}{\emph{Proceedings of the 9th ACM Conference on
  Recommender Systems}}. \bibinfo{pages}{155--162}.
\newblock


\bibitem[{City of Chicago}(2023)]%
        {dataset_chicagofood}
\bibfield{author}{\bibinfo{person}{{City of Chicago}}.}
  \bibinfo{year}{2023}\natexlab{}.
\newblock \bibinfo{title}{Food Inspections}.
\newblock \bibinfo{howpublished}{Chicago Data Portal,
  \url{https://data.cityofchicago.org/Health-Human-Services/Food-Inspections/4ijn-s7e5}}.
\newblock
\newblock
\shownote{Public Domain}.


\bibitem[Comanita(2024)]%
        {dataset_nationalgrid}
\bibfield{author}{\bibinfo{person}{Stefan Comanita}.}
  \bibinfo{year}{2024}\natexlab{}.
\newblock \bibinfo{title}{Hourly Electricity Consumption and Production}.
\newblock \bibinfo{howpublished}{Kaggle,
  \url{https://www.kaggle.com/datasets/stefancomanita/hourly-electricity-consumption-and-production}}.
\newblock
\newblock
\shownote{CC0-1.0}.


\bibitem[Daley and Vere-Jones(2003)]%
        {daley2003introduction}
\bibfield{author}{\bibinfo{person}{Daryl~J Daley} {and} \bibinfo{person}{David
  Vere-Jones}.} \bibinfo{year}{2003}\natexlab{}.
\newblock \showarticletitle{An introduction to the theory of point processes,
  volume 1: Elementary theory and methods}.
\newblock \bibinfo{journal}{\emph{Verlag New York Berlin Heidelberg: Springer}}
  (\bibinfo{year}{2003}).
\newblock


\bibitem[Daley and Vere-Jones(2008)]%
        {daley2008introduction}
\bibfield{author}{\bibinfo{person}{Daryl~J Daley} {and} \bibinfo{person}{David
  Vere-Jones}.} \bibinfo{year}{2008}\natexlab{}.
\newblock \bibinfo{booktitle}{\emph{An introduction to the theory of point
  processes: volume II: general theory and structure}}.
\newblock \bibinfo{publisher}{Springer}.
\newblock


\bibitem[Du et~al\mbox{.}(2016)]%
        {du2016recurrent}
\bibfield{author}{\bibinfo{person}{Nan Du}, \bibinfo{person}{Hanjun Dai},
  \bibinfo{person}{Rakshit Trivedi}, \bibinfo{person}{Utkarsh Upadhyay},
  \bibinfo{person}{Manuel Gomez-Rodriguez}, {and} \bibinfo{person}{Le Song}.}
  \bibinfo{year}{2016}\natexlab{}.
\newblock \showarticletitle{Recurrent marked temporal point processes:
  Embedding event history to vector}. In \bibinfo{booktitle}{\emph{Proceedings
  of the 22nd ACM SIGKDD International Conference on Knowledge Discovery and
  Data Mining}}. \bibinfo{pages}{1555--1564}.
\newblock


\bibitem[Evermann et~al\mbox{.}(2017)]%
        {evermann2017predicting}
\bibfield{author}{\bibinfo{person}{Joerg Evermann},
  \bibinfo{person}{Jana-Rebecca Rehse}, {and} \bibinfo{person}{Peter Fettke}.}
  \bibinfo{year}{2017}\natexlab{}.
\newblock \showarticletitle{Predicting Process Behaviour Using Deep Learning}.
\newblock \bibinfo{journal}{\emph{Decision Support Systems}}
  \bibinfo{volume}{100} (\bibinfo{year}{2017}), \bibinfo{pages}{129--140}.
\newblock
\href{https://doi.org/10.1016/j.dss.2017.04.003}{doi:\nolinkurl{10.1016/j.dss.2017.04.003}}


\bibitem[Gonzalez~Galtier(2024)]%
        {dataset_medicalappointments}
\bibfield{author}{\bibinfo{person}{Carolina Gonzalez~Galtier}.}
  \bibinfo{year}{2024}\natexlab{}.
\newblock \bibinfo{title}{Medical Appointment Scheduling System}.
\newblock \bibinfo{howpublished}{Kaggle,
  \url{https://www.kaggle.com/datasets/carogonzalezgaltier/medical-appointment-scheduling-system}}.
\newblock
\newblock
\shownote{CC BY 4.0}.


\bibitem[Gopalan et~al\mbox{.}(2014)]%
        {gopalan2014content}
\bibfield{author}{\bibinfo{person}{Prem Gopalan}, \bibinfo{person}{Laurent
  Charlin}, {and} \bibinfo{person}{David~M Blei}.}
  \bibinfo{year}{2014}\natexlab{}.
\newblock \showarticletitle{Content-based recommendations with Poisson
  factorization}.
\newblock \bibinfo{journal}{\emph{Advances in neural information processing
  systems}}  \bibinfo{volume}{27} (\bibinfo{year}{2014}).
\newblock


\bibitem[Gopalan et~al\mbox{.}(2015)]%
        {gopalan2015scalable}
\bibfield{author}{\bibinfo{person}{Prem Gopalan}, \bibinfo{person}{Jake~M
  Hofman}, {and} \bibinfo{person}{David~M Blei}.}
  \bibinfo{year}{2015}\natexlab{}.
\newblock \showarticletitle{Scalable recommendation with hierarchical poisson
  factorization.}. In \bibinfo{booktitle}{\emph{UAI}}.
  \bibinfo{pages}{326--335}.
\newblock


\bibitem[Grandell(1976)]%
        {grandell2006doubly}
\bibfield{author}{\bibinfo{person}{Jan Grandell}.}
  \bibinfo{year}{1976}\natexlab{}.
\newblock \bibinfo{booktitle}{\emph{Doubly stochastic {P}oisson processes}}.
  \bibinfo{series}{Lecture Notes in Mathematics}, Vol.~\bibinfo{volume}{529}.
\newblock \bibinfo{publisher}{Springer Berlin Heidelberg}.
\newblock
\showISBNx{978-3-540-07795-4}
\href{https://doi.org/10.1007/BFb0077758}{doi:\nolinkurl{10.1007/BFb0077758}}


\bibitem[Hawkes(1971)]%
        {hawkes1971spectra}
\bibfield{author}{\bibinfo{person}{Alan~G Hawkes}.}
  \bibinfo{year}{1971}\natexlab{}.
\newblock \showarticletitle{Spectra of some self-exciting and mutually exciting
  point processes}.
\newblock \bibinfo{journal}{\emph{Biometrika}} \bibinfo{volume}{58},
  \bibinfo{number}{1} (\bibinfo{year}{1971}), \bibinfo{pages}{83--90}.
\newblock


\bibitem[Jaegle et~al\mbox{.}(2021)]%
        {Jaegle21}
\bibfield{author}{\bibinfo{person}{Andrew Jaegle}, \bibinfo{person}{Felix
  Gimeno}, \bibinfo{person}{Andy Brock}, \bibinfo{person}{Oriol Vinyals},
  \bibinfo{person}{Andrew Zisserman}, {and} \bibinfo{person}{Jo{\~{a}}o
  Carreira}.} \bibinfo{year}{2021}\natexlab{}.
\newblock \showarticletitle{Perceiver: General Perception with Iterative
  Attention}. In \bibinfo{booktitle}{\emph{{ICML}}}
  \emph{(\bibinfo{series}{Proceedings of Machine Learning Research},
  Vol.~\bibinfo{volume}{139})}. \bibinfo{publisher}{{PMLR}},
  \bibinfo{pages}{4651--4664}.
\newblock


\bibitem[Kerrigan et~al\mbox{.}(2026)]%
        {kerrigan2026eventflow}
\bibfield{author}{\bibinfo{person}{Gavin Kerrigan}, \bibinfo{person}{Kai
  Nelson}, {and} \bibinfo{person}{Padhraic Smyth}.}
  \bibinfo{year}{2026}\natexlab{}.
\newblock \showarticletitle{Event{F}low: Forecasting Temporal Point Processes
  with Flow Matching}. In \bibinfo{booktitle}{\emph{The 29th International
  Conference on Artificial Intelligence and Statistics}}.
\newblock
\urldef\tempurl%
\url{https://openreview.net/forum?id=QXqKGOE2JW}
\showURL{%
\tempurl}


\bibitem[Lawless(1987)]%
        {lawless1987regression}
\bibfield{author}{\bibinfo{person}{Jerald~Franklin Lawless}.}
  \bibinfo{year}{1987}\natexlab{}.
\newblock \showarticletitle{Regression methods for Poisson process data}.
\newblock \bibinfo{journal}{\emph{J. Amer. Statist. Assoc.}}
  \bibinfo{volume}{82}, \bibinfo{number}{399} (\bibinfo{year}{1987}),
  \bibinfo{pages}{808--815}.
\newblock


\bibitem[Lin et~al\mbox{.}(2022)]%
        {lin2022exploring}
\bibfield{author}{\bibinfo{person}{Haitao Lin}, \bibinfo{person}{Lirong Wu},
  \bibinfo{person}{Guojiang Zhao}, \bibinfo{person}{Liu Pai}, {and}
  \bibinfo{person}{Stan~Z. Li}.} \bibinfo{year}{2022}\natexlab{}.
\newblock \showarticletitle{Exploring Generative Neural Temporal Point
  Process}.
\newblock \bibinfo{journal}{\emph{Transactions on Machine Learning Research}}
  (\bibinfo{year}{2022}).
\newblock
\showISSN{2835-8856}
\urldef\tempurl%
\url{https://openreview.net/forum?id=NPfS5N3jbL}
\showURL{%
\tempurl}


\bibitem[L{\"u}dke et~al\mbox{.}(2023)]%
        {ludke2023add}
\bibfield{author}{\bibinfo{person}{David L{\"u}dke}, \bibinfo{person}{Marin
  Bilo{\v{s}}}, \bibinfo{person}{Oleksandr Shchur}, \bibinfo{person}{Marten
  Lienen}, {and} \bibinfo{person}{Stephan G{\"u}nnemann}.}
  \bibinfo{year}{2023}\natexlab{}.
\newblock \showarticletitle{Add and thin: Diffusion for temporal point
  processes}.
\newblock \bibinfo{journal}{\emph{Advances in Neural Information Processing
  Systems}}  \bibinfo{volume}{36} (\bibinfo{year}{2023}),
  \bibinfo{pages}{56784--56801}.
\newblock


\bibitem[L{\"u}dke et~al\mbox{.}(2026)]%
        {ludke2026editbased}
\bibfield{author}{\bibinfo{person}{David L{\"u}dke}, \bibinfo{person}{Marten
  Lienen}, \bibinfo{person}{Marcel Kollovieh}, {and} \bibinfo{person}{Stephan
  G{\"u}nnemann}.} \bibinfo{year}{2026}\natexlab{}.
\newblock \showarticletitle{Edit-Based Flow Matching for Temporal Point
  Processes}. In \bibinfo{booktitle}{\emph{The Fourteenth International
  Conference on Learning Representations}}.
\newblock
\urldef\tempurl%
\url{https://openreview.net/forum?id=FNf9IV1P2L}
\showURL{%
\tempurl}


\bibitem[Mei and Eisner(2017)]%
        {mei2017neural}
\bibfield{author}{\bibinfo{person}{Hongyuan Mei} {and} \bibinfo{person}{Jason~M
  Eisner}.} \bibinfo{year}{2017}\natexlab{}.
\newblock \showarticletitle{The neural hawkes process: A neurally
  self-modulating multivariate point process}. In
  \bibinfo{booktitle}{\emph{Advances in Neural Information Processing
  Systems}}. \bibinfo{pages}{6754--6764}.
\newblock


\bibitem[Meng et~al\mbox{.}(2024)]%
        {meng2024transfeattpp}
\bibfield{author}{\bibinfo{person}{Zizhuo Meng}, \bibinfo{person}{Boyu Li},
  \bibinfo{person}{Xuhui Fan}, \bibinfo{person}{Zhidong Li},
  \bibinfo{person}{Yang Wang}, \bibinfo{person}{Fang Chen}, {and}
  \bibinfo{person}{Feng Zhou}.} \bibinfo{year}{2024}\natexlab{}.
\newblock \showarticletitle{TransFeat-TPP: An Interpretable Deep Covariate
  Temporal Point Processes}. In \bibinfo{booktitle}{\emph{{ECAI}}}
  \emph{(\bibinfo{series}{Frontiers in Artificial Intelligence and
  Applications}, Vol.~\bibinfo{volume}{392})}. \bibinfo{publisher}{{IOS}
  Press}, \bibinfo{pages}{810--817}.
\newblock


\bibitem[Navarin et~al\mbox{.}(2017)]%
        {navarin2017lstm}
\bibfield{author}{\bibinfo{person}{Nicol{\`o} Navarin},
  \bibinfo{person}{Beatrice Vincenzi}, \bibinfo{person}{Mirko Polato}, {and}
  \bibinfo{person}{Alessandro Sperduti}.} \bibinfo{year}{2017}\natexlab{}.
\newblock \showarticletitle{{LSTM} Networks for Data-Aware Remaining Time
  Prediction of Business Process Instances}. In \bibinfo{booktitle}{\emph{2017
  IEEE Symposium Series on Computational Intelligence (SSCI)}}.
  \bibinfo{publisher}{IEEE}, \bibinfo{pages}{1--7}.
\newblock
\href{https://doi.org/10.1109/SSCI.2017.8285184}{doi:\nolinkurl{10.1109/SSCI.2017.8285184}}


\bibitem[Omi et~al\mbox{.}(2019)]%
        {omi2019fully}
\bibfield{author}{\bibinfo{person}{Takahiro Omi}, \bibinfo{person}{Kazuyuki
  Aihara}, {et~al\mbox{.}}} \bibinfo{year}{2019}\natexlab{}.
\newblock \showarticletitle{Fully neural network based model for general
  temporal point processes}. In \bibinfo{booktitle}{\emph{Advances in Neural
  Information Processing Systems}}. \bibinfo{pages}{2122--2132}.
\newblock


\bibitem[Pan et~al\mbox{.}(2021)]%
        {pan2021self}
\bibfield{author}{\bibinfo{person}{Zhimeng Pan}, \bibinfo{person}{Zheng Wang},
  \bibinfo{person}{Jeff~M Phillips}, {and} \bibinfo{person}{Shandian Zhe}.}
  \bibinfo{year}{2021}\natexlab{}.
\newblock \showarticletitle{Self-adaptable point processes with nonparametric
  time decays}.
\newblock \bibinfo{journal}{\emph{Advances in Neural Information Processing
  Systems}}  \bibinfo{volume}{34} (\bibinfo{year}{2021}),
  \bibinfo{pages}{4594--4606}.
\newblock


\bibitem[Pan et~al\mbox{.}(2020)]%
        {pan2020scalable}
\bibfield{author}{\bibinfo{person}{Zhimeng Pan}, \bibinfo{person}{Zheng Wang},
  {and} \bibinfo{person}{Shandian Zhe}.} \bibinfo{year}{2020}\natexlab{}.
\newblock \showarticletitle{Scalable nonparametric factorization for high-order
  interaction events}. In \bibinfo{booktitle}{\emph{International Conference on
  Artificial Intelligence and Statistics}}. PMLR, \bibinfo{pages}{4325--4335}.
\newblock


\bibitem[Pegoraro et~al\mbox{.}(2021)]%
        {pegoraro2021text}
\bibfield{author}{\bibinfo{person}{Marco Pegoraro},
  \bibinfo{person}{Merih~Seran Uysal}, \bibinfo{person}{David~Benedikt Georgi},
  {and} \bibinfo{person}{Wil M.~P. van~der Aalst}.}
  \bibinfo{year}{2021}\natexlab{}.
\newblock \showarticletitle{Text-Aware Predictive Monitoring of Business
  Processes}. In \bibinfo{booktitle}{\emph{Business Information Systems}}.
  \bibinfo{publisher}{TIB Open Publishing}, \bibinfo{pages}{221--232}.
\newblock
\href{https://doi.org/10.52825/bis.v1i.62}{doi:\nolinkurl{10.52825/bis.v1i.62}}


\bibitem[Shchur et~al\mbox{.}(2020)]%
        {Shchur2020IntensityFree}
\bibfield{author}{\bibinfo{person}{Oleksandr Shchur}, \bibinfo{person}{Marin
  Bilos}, {and} \bibinfo{person}{Stephan G{\"{u}}nnemann}.}
  \bibinfo{year}{2020}\natexlab{}.
\newblock \showarticletitle{Intensity-Free Learning of Temporal Point
  Processes}. In \bibinfo{booktitle}{\emph{{ICLR}}}.
  \bibinfo{publisher}{OpenReview.net}.
\newblock


\bibitem[Sill(1997)]%
        {sill1997monotonic}
\bibfield{author}{\bibinfo{person}{Joseph Sill}.}
  \bibinfo{year}{1997}\natexlab{}.
\newblock \showarticletitle{Monotonic networks}.
\newblock \bibinfo{journal}{\emph{Advances in neural information processing
  systems}}  \bibinfo{volume}{10} (\bibinfo{year}{1997}).
\newblock


\bibitem[Tax et~al\mbox{.}(2017)]%
        {tax2017predictive}
\bibfield{author}{\bibinfo{person}{Niek Tax}, \bibinfo{person}{Ilya Verenich},
  \bibinfo{person}{Marcello La~Rosa}, {and} \bibinfo{person}{Marlon Dumas}.}
  \bibinfo{year}{2017}\natexlab{}.
\newblock \showarticletitle{Predictive Business Process Monitoring with {LSTM}
  Neural Networks}. In \bibinfo{booktitle}{\emph{Advanced Information Systems
  Engineering}} \emph{(\bibinfo{series}{Lecture Notes in Computer Science},
  Vol.~\bibinfo{volume}{10253})}. \bibinfo{publisher}{Springer},
  \bibinfo{pages}{477--492}.
\newblock
\href{https://doi.org/10.1007/978-3-319-59536-8_30}{doi:\nolinkurl{10.1007/978-3-319-59536-8_30}}


\bibitem[van~der Aalst(2016)]%
        {vanderaalst2016process}
\bibfield{author}{\bibinfo{person}{Wil M.~P. van~der Aalst}.}
  \bibinfo{year}{2016}\natexlab{}.
\newblock \bibinfo{booktitle}{\emph{Process Mining: Data Science in Action}
  (\bibinfo{edition}{2} ed.)}.
\newblock \bibinfo{publisher}{Springer}, \bibinfo{address}{Berlin, Heidelberg}.
\newblock
\showISBNx{978-3-662-49850-7}
\href{https://doi.org/10.1007/978-3-662-49851-4}{doi:\nolinkurl{10.1007/978-3-662-49851-4}}


\bibitem[Wang et~al\mbox{.}(2023)]%
        {wang2023mitfm}
\bibfield{author}{\bibinfo{person}{Jiaxing Wang}, \bibinfo{person}{Chengliang
  Lu}, \bibinfo{person}{Bin Cao}, {and} \bibinfo{person}{Jing Fan}.}
  \bibinfo{year}{2023}\natexlab{}.
\newblock \showarticletitle{{MiTFM}: A Multi-View Information Fusion Method
  Based on Transformer for Next Activity Prediction of Business Processes}. In
  \bibinfo{booktitle}{\emph{Proceedings of the 14th Asia-Pacific Symposium on
  Internetware}}. \bibinfo{publisher}{ACM}, \bibinfo{pages}{281--291}.
\newblock
\href{https://doi.org/10.1145/3609437.3609442}{doi:\nolinkurl{10.1145/3609437.3609442}}


\bibitem[Wang et~al\mbox{.}(2017)]%
        {wang2017predicting}
\bibfield{author}{\bibinfo{person}{Yichen Wang}, \bibinfo{person}{Xiaojing Ye},
  \bibinfo{person}{Hongyuan Zha}, {and} \bibinfo{person}{Le Song}.}
  \bibinfo{year}{2017}\natexlab{}.
\newblock \showarticletitle{Predicting user activity level in point processes
  with mass transport equation}. In \bibinfo{booktitle}{\emph{Advances in
  Neural Information Processing Systems}}. \bibinfo{pages}{1644--1654}.
\newblock


\bibitem[Xu et~al\mbox{.}(2018)]%
        {xu2018benefits}
\bibfield{author}{\bibinfo{person}{Hongteng Xu}, \bibinfo{person}{Dixin Luo},
  \bibinfo{person}{Xu Chen}, {and} \bibinfo{person}{Lawrence Carin}.}
  \bibinfo{year}{2018}\natexlab{}.
\newblock \showarticletitle{Benefits from superposed hawkes processes}. In
  \bibinfo{booktitle}{\emph{International Conference on Artificial Intelligence
  and Statistics}}. PMLR, \bibinfo{pages}{623--631}.
\newblock


\bibitem[Xue et~al\mbox{.}(2024)]%
        {xueeasytpp}
\bibfield{author}{\bibinfo{person}{Siqiao Xue}, \bibinfo{person}{Xiaoming Shi},
  \bibinfo{person}{Zhixuan Chu}, \bibinfo{person}{Yan Wang},
  \bibinfo{person}{Hongyan Hao}, \bibinfo{person}{Fan Zhou},
  \bibinfo{person}{Caigao JIANG}, \bibinfo{person}{Chen Pan},
  \bibinfo{person}{James~Y Zhang}, \bibinfo{person}{Qingsong Wen},
  {et~al\mbox{.}}} \bibinfo{year}{2024}\natexlab{}.
\newblock \showarticletitle{EasyTPP: Towards Open Benchmarking Temporal Point
  Processes}. In \bibinfo{booktitle}{\emph{The Twelfth International Conference
  on Learning Representations}}.
\newblock


\bibitem[Yang et~al\mbox{.}(2022)]%
        {yang2022transformer}
\bibfield{author}{\bibinfo{person}{Chenghao Yang}, \bibinfo{person}{Hongyuan
  Mei}, {and} \bibinfo{person}{Jason Eisner}.} \bibinfo{year}{2022}\natexlab{}.
\newblock \showarticletitle{Transformer Embeddings of Irregularly Spaced Events
  and Their Participants}. In \bibinfo{booktitle}{\emph{Proceedings of the
  Tenth International Conference on Learning Representations (ICLR)}}.
\newblock


\bibitem[Yang et~al\mbox{.}(2017)]%
        {yang2017decoupling}
\bibfield{author}{\bibinfo{person}{Jiasen Yang}, \bibinfo{person}{Vinayak~A
  Rao}, {and} \bibinfo{person}{Jennifer Neville}.}
  \bibinfo{year}{2017}\natexlab{}.
\newblock \showarticletitle{Decoupling Homophily and Reciprocity with Latent
  Space Network Models.}. In \bibinfo{booktitle}{\emph{UAI}}.
\newblock


\bibitem[Yuan and Fang(2025)]%
        {yuan2025residual}
\bibfield{author}{\bibinfo{person}{Ruoxin Yuan} {and} \bibinfo{person}{Guanhua
  Fang}.} \bibinfo{year}{2025}\natexlab{}.
\newblock \showarticletitle{Residual {TPP}: A Unified Lightweight Approach for
  Event Stream Data Analysis}. In \bibinfo{booktitle}{\emph{Forty-second
  International Conference on Machine Learning}}.
\newblock


\bibitem[Zhang et~al\mbox{.}(2020)]%
        {zhang2020self}
\bibfield{author}{\bibinfo{person}{Qiang Zhang}, \bibinfo{person}{Aldo Lipani},
  \bibinfo{person}{Omer Kirnap}, {and} \bibinfo{person}{Emine Yilmaz}.}
  \bibinfo{year}{2020}\natexlab{}.
\newblock \showarticletitle{Self-attentive hawkes process}. In
  \bibinfo{booktitle}{\emph{International Conference on Machine Learning}}.
  PMLR, \bibinfo{pages}{11183--11193}.
\newblock


\bibitem[Zhe and Du(2018)]%
        {zhe2018stochastic}
\bibfield{author}{\bibinfo{person}{Shandian Zhe} {and} \bibinfo{person}{Yishuai
  Du}.} \bibinfo{year}{2018}\natexlab{}.
\newblock \showarticletitle{Stochastic Nonparametric Event-Tensor
  Decomposition}. In \bibinfo{booktitle}{\emph{Advances in Neural Information
  Processing Systems}}. \bibinfo{pages}{6856--6866}.
\newblock


\bibitem[Zuo et~al\mbox{.}(2020)]%
        {zuo2020transformer}
\bibfield{author}{\bibinfo{person}{Simiao Zuo}, \bibinfo{person}{Haoming
  Jiang}, \bibinfo{person}{Zichong Li}, \bibinfo{person}{Tuo Zhao}, {and}
  \bibinfo{person}{Hongyuan Zha}.} \bibinfo{year}{2020}\natexlab{}.
\newblock \showarticletitle{Transformer hawkes process}. In
  \bibinfo{booktitle}{\emph{International Conference on Machine Learning}}.
  PMLR, \bibinfo{pages}{11692--11702}.
\newblock


\end{thebibliography}

\end{document}